\documentclass{article}

\usepackage{microtype}
\usepackage{amssymb}
\usepackage{graphicx}
\usepackage{subcaption}
\usepackage{mathrsfs}
\usepackage{color}
\usepackage{booktabs} 
\usepackage{breqn}
\usepackage{bm, bbm}
\usepackage{hyperref}
\usepackage{enumitem}

\usepackage[accepted]{icml2024}

\usepackage{amsmath}
\usepackage{amssymb}
\usepackage{mathtools}
\usepackage{amsthm}
\def\calF{\mathcal{F}}
\def\logit{\mathrm{logit}}
\def\IIF{\mathsf{IF}}
\def\EIF{\mathsf{EIF}}
\usepackage[capitalize,noabbrev]{cleveref}

\def\dnase{{\mathsf{DNA\mbox{-}SE}}}

\def\calN{\mathcal{N}}

\def\tanh{\mathsf{tanh}}

\def\eff{\mathsf{eff}}

\def\opt{\mathsf{opt}}

\def\sfb{\mathsf{b}}

\def\sfb{\mathsf{b}}

\def\bfX{\mathbf{X}}
\def\bfx{\mathbf{x}}
\def\var{\mathrm{var}}

\def\expit{\mathrm{expit}}

\def\sfs{\mathsf{s}}
\theoremstyle{plain}
\newtheorem{theorem}{Theorem}[section]

\theoremstyle{definition}

\newtheorem{example}[theorem]{Example}
\theoremstyle{remark}
\newtheorem{remark}[theorem]{Remark}

\def\diff{\mathrm{d}}
\def\P{\mathrm{P}}
\def\bbeta{\boldsymbol{\beta}}
\def\E{\mathrm{E}}

\def\hat{\widehat}
\def\tilde{\widetilde}
\def\p{\mathrm{p}}
\usepackage[textsize=tiny]{todonotes}
\def\st{\mathrm{s.t.}}
\def\bbR{\mathbb{R}}

\icmltitlerunning{DNN-Assisted Semiparametric Estimation}

\begin{document}

\twocolumn[
\icmltitle{DNA-SE: Towards Deep Neural-Nets Assisted Semiparametric Estimation}



\icmlsetsymbol{equal}{*}
\icmlsetsymbol{co}{$\dag$}

\begin{icmlauthorlist}
\icmlauthor{Qinshuo Liu}{1}
\icmlauthor{Zixin Wang}{2}
\icmlauthor{Xi-An Li}{3}
\icmlauthor{Xinyao Ji}{}
\icmlauthor{Lei Zhang}{3}
\icmlauthor{Lin Liu}{co,3,4}
\icmlauthor{Zhonghua Liu}{co,5}
\end{icmlauthorlist}

\icmlaffiliation{1}{Department of Statistics and Actuarial Science, The University of Hong Kong, Hong Kong SAR, China}
\icmlaffiliation{2}{School of Life Sciences, Shanghai Jiao Tong University, Shanghai, China}
\icmlaffiliation{3}{Institute of Natural Sciences, MOE-LSC, School of Mathematical Sciences, CMA-Shanghai, Shanghai Jiao Tong University, Shanghai, China}
\icmlaffiliation{4}{SJTU-Yale Joint Center for Biostatistics and Data Science, Shanghai Jiao Tong University, Shanghai, China}
\icmlaffiliation{5}{Department of Biostatistics, Columbia University, New York, NY, USA}
\icmlcorrespondingauthor{Lin Liu}{linliu@sjtu.edu.cn}
\icmlcorrespondingauthor{Zhonghua Liu}{zl2509@cumc.columbia.edu}

\icmlkeywords{semiparametric statistics, missing data, causal inference, Fredholm integral equations of the second kind, bi-level optimization, deep learning, AI for science}

\vskip 0.3in
]



\printAffiliationsAndNotice{The two corresponding authors are alphabetically ordered and contributed equally to this project.} 

\begin{abstract}

Semiparametric statistics play a pivotal role in a wide range of domains, including but not limited to missing data, causal inference, and transfer learning, to name a few. In many settings, semiparametric theory leads to (nearly) statistically optimal procedures that yet involve numerically solving Fredholm integral equations of the second kind. Traditional numerical methods, such as polynomial or spline approximations, are difficult to scale to multi-dimensional problems. Alternatively, statisticians may choose to approximate the original integral equations by ones with closed-form solutions, resulting in computationally more efficient, but statistically suboptimal or even incorrect procedures. To bridge this gap, we propose a novel framework by formulating the semiparametric estimation problem as a bi-level optimization problem; and then we develop a scalable algorithm called \underline{D}eep \underline{N}eural-Nets \underline{A}ssisted \underline{S}emiparametric \underline{E}stimation ($\dnase$) by leveraging the universal approximation property of Deep Neural-Nets (DNN) to streamline semiparametric procedures. Through extensive numerical experiments and a real data analysis, we demonstrate the numerical and statistical advantages of $\dnase$ over traditional methods. To the best of our knowledge, we are the first to bring DNN into semiparametric statistics as a numerical solver of integral equations in our proposed general framework. 
\end{abstract}

\section{Introduction}
\label{Introduction_chap}

Deep Neural-Nets (DNN) have made unprecedented progress in a variety of scientific and industrial problems, such as image classification \citep{krizhevsky2012imagenet}, natural language processing \citep{vaswani2017attention}, and scientific computing \citep{e2018deep}. Recently, the statistical literature has also garnered increasing interests in deploying DNN to solve statistical problems that baffled traditional methods, for example, nonparametric regression \citep{chen1999improved, suzuki2019adaptivity, schmidt2020nonparametric, padilla2022quantile} and nonparametric density estimation \citep{liu2021density, liang2021well}. A common theme in these applications is to view DNN as an alternative method to approximate infinite-dimensional nonparametric models, conventionally done by splines, polynomials, etc. \citep{gine2016mathematical}.
 
Interpretability is at the core of many statistical problems. A conceptually appealing statistical paradigm is semiparametric statistics (including semiparametric statistical models, methods and related theory). In semiparametric statistics, one models the part of the Data Generating Process (DGP) of scientific interest with low-dimensional parametric models to improve interpretability, while leaving the remaining part (termed as nuisance parameters) either completely unspecified or modeled by infinite-dimensional nonparametric models to avoid unnecessary model misspecification bias \citep{newey1990semiparametric, bickel1998efficient}. As a result, semiparametric statistics have been very popular in problems calling for both interpretability and robustness against model misspecification, such as missing data \citep{robins1994estimation}, causal inference \citep{van2003unified}, and more recently, transfer learning \citep{qiu2023efficient, tian2023elsa}. It is natural to employ DNN to estimate the infinite-dimensional nuisance parameters in semiparametric models, an extension of the aforementioned idea from nonparametric statistics. In fact, this is the route that most recent semiparametric statistics literature has taken \citep{farrell2021deep, shi2021double, zhong2022deep, xu2022deepmed, kompa2022deep, chernozhukov2022riesznet, chen2023efficient, chen2024causal}. 

In this paper, we take a different path and instead propose to treat DNN as a numerical solver for semiparametric statistics. Here is our motivation. All the above cited works on semiparametric statistics are special cases, where the statistically optimal (i.e. semiparametric efficient) estimator for the parameter of scientific interest has closed-form. In general, however, semiparametric theory can lead to nearly statistically optimal estimators that depend on the solution of Fredholm integral equations of the second kind\footnote{In the sequel, we will simply call Fredholm integral equations of the second kind as integral equations to ease exposition.} \citep{bickel1998efficient}. These integral equations often: (1) do not have closed-form solutions; and (2) depend on the unknown parameter of interest, calling for complicated iterative methods \citep{robins1994estimation}. As shown in our numerical experiments in Section \ref{sec:sim}, using traditional methods such as polynomials/splines \citep{bellour2016two} or approximately inverting the integral kernel \citep{atkinson1967numerical} to approximate the solution does not always lead to optimal or satisfactory estimators in practice. In certain cases, statisticians may derive alternative estimators with closed-forms, but this strategy generally leads to statistically suboptimal or sometimes even inconsistent estimators \citep{zhao2022versatile, liu2021efficient, yang2022semiparametric}.

To bridge this gap, inspired by recent works using DNN as a new-generation numerical method for scientific computing \citep{e2018deep, raissi2019physics, li2020multi}, we propose a novel framework by formulating the semiparametric estimation problem as a bi-level optimization problem; and then we develop a scalable algorithm called \underline{D}eep \underline{N}eural-Nets \underline{A}ssisted \underline{S}emiparametric \underline{E}stimation ($\dnase$). In a nutshell, $\dnase$ uses multi-layer feed-forward neural networks to approximate the solution of the integral equations and then employs the alternating gradient descent algorithm to solve the integral equations and estimate the parameters of interest simultaneously (see Section \ref{sec:algor}). We summarize our main contributions below.

\subsection*{Main Contributions}
\allowdisplaybreaks
\begin{itemize}
\setlength\itemsep{0.5pt}
    \item We develop a new DNN-based algorithm called $\dnase$ that streamlines solving the integral equations and estimating the parameter of interest simultaneously, by (1) formulating the semiparametric estimation problem as a bi-level optimization problem minimizing loss functions with a ``nested'' structure; and (2) solving this bi-level programming via Alternating Gradient Descent algorithms. To our knowledge, our paper is the first that employs DNN to numerically solve integral equations that arise in semiparametric problems and draws connections between semiparametric statistics and bi-level optimization \citep{chen2021closing}. 
    \item We investigate the empirical performance of $\dnase$, together with the issue of hyperparameter tuning through numerical experiments and a real data analysis, which demonstrates $\dnase$ as a promising practical tool for estimating parameters in semiparametric models,  outperforming traditional methods (e.g. polynomials) \cite{atkinson1967numerical, ren1999simple}.
    \item We have wrapped $\dnase$ into a python package, which is accessible via \href{https://github.com/liuqs111/DNA-SE}{\underline{this link}}. Semiparametric methods have been viewed as a relatively difficult subject  for some practitioners \citep{ hines2022demystifying}. We envision that $\dnase$ and its future version, together with other related open-source software \citep{lee2023ananke}, can help popularize semiparametric methods in artificial intelligence and other related scientific disciplines \citep{pan2022knowledge, min2023nonparametric, wang2023scientific}.
\end{itemize}

\subsection*{Other Related Works}

To the best of our knowledge, our work is the first attempt to bring DNN into semiparametric statistics as a numerical solver for integral equations. Recently, there have been a few works using DNN to solve a variety of integral equations \citep{guan2022solving, zappala2023neural}. In semiparametric statistics, however, solving integral equations is only a means to an end (parameter estimation), and more importantly, the integral equations appeared in semiparametric problems often depend on the unknown parameters to be estimated, except in some special cases (e.g. Example \ref{eg:shift}). Therefore these prior works cannot be directly applied to semiparametric statistics in the most general form.


\subsection*{Organization of the Paper}
The rest of our paper is organized as follows. In Section \ref{sec:algor}, we briefly review semiparametric statistics, describe the problem, and formulate parameter estimation under semiparametric models as a generic bi-level optimization problem, that minimizes a set of nested loss functions. The $\dnase$ algorithm is then proposed to solve this type of problems. Three concrete examples from missing data, causal inference and transfer learning are used to illustrate our algorithm in Section \ref{sec:examples}. As a proof-of-concept, numerical experiments and a real data analysis are, respectively, conducted in Section \ref{sec:sim} and Section \ref{sec:data}. We conclude our article and discuss a few future research avenues in Section \ref{sec:conclude}. Minor details are deferred to the Appendix.

\subsection*{Notation}

Before proceeding, we collect some notation used throughout. Capital letters are reserved for random variables and lower-case letters are used for their realized values. The true data generating distribution is denoted as $\P$, the probability density function (p.d.f.) or the probability mass function (p.m.f.) of which is denoted as $\p$. 
Throughout the paper, $\bbeta$ denotes the parameter of scientific interest (possibly vector-valued) and $\bm{\eta}$ the nuisance parameter. We let $\calF_{L, K}$ be the class of fully-connected feedforward DNN with depth $L$ and width $K$, whose intput and output dimensions will be clear later. This is the type of DNN that will be considered in this paper, although other more complex architectures (e.g. Transformers \citep{vaswani2017attention}) may also be explored in the future. $\bm{\omega}$ is reserved for the neuron weight parameters. Finally, we denote $\{O_i\}_{i = 1}^N$ as the $N$ i.i.d. observed data drawn from $\P$. In general, $O$ has several components, e.g. $O = (O_1, O_2)$. We denote the conditional distribution of $O_{2} | O_{1}$ as $\P_{O_{2} | O_{1}}$.

\section{Main Results: A New Framework and the  $\dnase$ Algorithm}
\label{sec:algor}

\subsection{Problem Setup of Semiparametric Estimation}
\label{sec:setup}

We begin by describing the generic setup for parameter estimation that arises ubiquitously from semiparametric problems. Concrete examples that facilitate understanding the high-level description here are provided later in Section \ref{sec:examples}. Let $\bbeta \equiv \bbeta (\P) \in \mathbb{R}^{q}$ denote the parameter of scientific interest, with a fixed dimension $q > 0$. $\bm{\eta} \equiv \bm{\eta} (\P)$ denotes the nuisance parameter, which can be infinite-dimensional. Semiparametric theory \citep{bickel1998efficient, van2002part} often leads to the following system of moment estimating equations that the true $\bbeta^{\ast}$ must satisfy:
\allowdisplaybreaks
\begin{equation}
\small
\begin{split}
\E \left[ \psi (O; \sfb^{\ast} (\cdot), \bbeta^\ast) \right] & = 0, \text{where $\sfb^{\ast}$ solves} \\
\int K(\mathbf{s}, \mathbf{t}, O; \bbeta^\ast) \mathsf{b}^{\ast} (\mathbf{s}, O) \diff \mathbf{s} & = C(\mathbf{t}, O; \bbeta^{\ast})+\sfb^{\ast} (\mathbf{t}, O).
\end{split}
\label{general form sum}
\end{equation}
Here $\psi \in \mathbb{R}^{q}$ denotes the estimating functions of the same dimension as $\bbeta$ and $\sfb^{\ast} (\cdot)$ with output dimension $q$ is the solution to the system of $q$ integral equations in the second line of \eqref{general form sum}. $K(\cdot)$ is a kernel function known up to the parameter of interest $\bbeta$ and the nuisance parameter $\bm{\eta}$, $C (\cdot)$ is a known vector-valued function of the same dimension as $\bbeta$, also known up to $\bbeta$ and $\bm{\eta}$. To avoid clutter, we omit the dependence on $\bm{\eta}$ of these functions throughout. $\mathbf{s}$ and $\mathbf{t}$ denote variables lying in the same sample space as (possibly parts of) the data $O$ and they will be made explicit in the concrete examples later. In general the solution $\sfb^{\ast} (\cdot)$ implicitly depends on the unknown $\bbeta$, but there are exceptions (e.g. Example \ref{eg:shift} in Section \ref{sec:examples}). In certain cases, the nuisance parameter $\bm{\eta}$ can even be set arbitrarily, and semiparametric theory can help eliminate the dependence of $\hat{\bbeta}$ on $\bm{\eta}$ \textit{entirely}: e.g., see Examples \ref{eg:mnar} and \ref{eg:sensitivity}. An estimator $\hat{\bbeta}$ of $\bbeta^\ast$ can be obtained by solving the empirical analogue of \eqref{general form sum} \citep{van2023weak}.

\subsection{Formulating Semiparametric Estimation as a Bi-level Optimization Problem}
\label{sec:loss}
 Here, we propose a new framework by formulating the estimating equations \eqref{general form sum} into a bi-level optimization problem, which is amenable to being solved by modern DNN training algorithms. Specifically, we propose to minimize the following nested loss function to simultaneously solve the integral equation and estimate the parameter of interest $\bbeta$:
\begin{equation}
\label{loss_func}
\small
\begin{split}
& \hat{\bbeta} \coloneqq \arg\min_{\bbeta \in \bbR^{q}} \mathcal{L}_{\psi} (\bbeta; \hat{\sfb}) \ \st \ \hat{\sfb} \coloneqq \arg\min_{\sfb \in \calF} \mathcal{L}_{K} (\sfb), \text{where} \\
& \mathcal{L}_{\psi} (\bbeta; \sfb) \coloneqq \left\{ \frac{1}{N} \sum_{i = 1}^{N} \psi (O_{i}; \bbeta, \sfb) \right\}^{2} \text{ and} \\
& \mathcal{L}_{K} (\sfb) \coloneqq \frac{1}{N} \sum_{i = 1}^{N} \int_{\mathbf{t}} \left( \begin{array}{c}
\int K (\mathbf{s}, \mathbf{t}, O_{i}; \bbeta) \sfb (\mathbf{s}, O_{i}) \diff \mathbf{s} \\
- \, C (\mathbf{t}, O_{i}; \bbeta) - \sfb (\mathbf{t}, O_{i})
\end{array} \right)^{2} \diff \nu (\mathbf{t}),
\end{split}
\end{equation}
where $\nu (\cdot)$ is an easy-to-sample probability distribution over the sample space of $\mathbf{t}$ (default: uniform over a sufficiently large but bounded domain). It is noteworthy that the loss function \eqref{loss_func} converges to its population version (see Appendix \ref{app:loss}) and $K, C$ may depend on the estimate $\hat{\bm{\eta}}$ of $\bm{\eta}$ but again we do not make it explicit.

Note that the loss function \eqref{loss_func} has a distinctive nested structure, in the sense that the outer loss $\mathcal{L}_{\psi}$ depends on the solution to the inner loss $\mathcal{L}_{K}$. Here $\calF$ is some hypothesis class for approximating $\sfb (\cdot)$; in this paper, we choose $\calF \equiv \calF_{L, K}$, the class of fully-connected feedforward DNN. Then $\sfb (\cdot)$ can be parameterized by the neuron weights as $\sfb (\cdot; \bm{\omega})$ and thus $\hat{\sfb} (\cdot) \equiv \sfb (\cdot; \hat{\bm{\omega}})$, where $\hat{\bm{\omega}}$ denotes the neuron weights trained by minimizing \eqref{loss_func}.

In general, the integrals involved in \eqref{loss_func} need to be evaluated numerically. Here we use Monte Carlo integration \citep{liu2001monte} by drawing $J_{1}$ samples from $\nu$ to approximate the outer integral and $J_{2}$ samples uniformly from the space of $\mathbf{s}$ to approximate the inner integral of $\mathcal{L}_{K} (\sfb)$, although other methods can also be used in principle. For example, for fixed $\mathbf{t}$ and $\bm{\omega}$, the inner integral in $\mathcal{L}_{K} (\sfb)$ is approximated by:
\begin{equation*}
\small
\begin{aligned}
& \int K \left(\mathbf{s}, \mathbf{t}, O_i; \bbeta \right) \sfb\left(\mathbf{s},O_i;\bm{\omega} \right) \diff \mathbf{s} \\ & \quad \quad \quad \quad \approx 
\frac{\mathrm{vol}_{\mathbf{s}}}{J_2}\sum_{j=1}^{J_2} K\left(\mathbf{s}_j, \mathbf{t}_l,O_i; \bbeta\right) \sfb \left(\mathbf{s}_j,O_i;  \bm{\omega}\right),
\end{aligned}
\end{equation*}
where $\mathrm{vol}_{\mathbf{s}}$ is the volume of the sample space of $\mathbf{s}$, $\mathrm{vol}_{\mathbf{s}} = \int _\Omega \diff \mathbf{s}$, $\Omega$ is the domain of $\mathbf{s}$. As $J_{1}, J_{2} \rightarrow \infty$, the Monte Carlo approximations converge to the true integrals in probability by the law of large numbers. We recommend that they be set as large as the computational resource allows to minimize the numerical approximation error.

\subsection {The Alternating Gradient-Descent Algorithm}
\label{sec:opt}

Since minimizing the loss function \eqref{pop loss} constitutes a bi-level optimization problem, we adopt the Alternating Gradient Descent (GD) and its  extensions (e.g. its stochastic/adam version), one of the most natural algorithmic choices for bi-level programming \citep{finn2017model, chen2021closing}.

The algorithm is described in Algorithm \ref{alg:fred}. At the same time, the flow chart and network structure of the whole process are shown in Appendix \ref{app:figs}. We highlight several key features. First, the algorithm alternates between two types of GD updates -- one for minimizing the outer loss $\hat{\mathcal{L}}_{\psi}$ and the other for minimizing the inner loss $\hat{\mathcal{L}}_{K}$. Second, it involves a hyperparameter $\gamma$, the ``alternating frequency'', determining the number of GD updates for the inner loss before switching to GD updates for the outer loss.

We end this section with a few remarks.
\begin{remark}
Although we ground our work in the context of semiparametric estimation, this new framework can go beyond semiparametric statistics and be adapted to models subject to general (conditional) moment restrictions, commonly encountered in the statistics and econometrics literature \citep{ai2003efficient, chen2018overidentification, cui2023semiparametric}.
\end{remark}
\begin{remark}
In later numerical experiments (Section \ref{sec:sim}), we adopt the Alternating Stochastic GD \citep{wilson2017marginal, zhou2020towards} to solve the optimization problem \eqref{loss_func}. However, we point out that the component $\mathcal{L}_{\psi}$ corresponding to the parameter of interest $\bbeta$ in \eqref{loss_func} often has much fewer degrees-of-freedom than $\mathcal{L}_{K}$ for solving the integral equations. Thus in practice, one could also exploit optimization algorithms with greater computational complexities to minimize $\mathcal{L}_{\psi}$, to ensure a convergence to the actual global minimum. Furthermore, since algorithms for bi-level optimization problems are generally difficult to analyze \citep{chen2021closing}, we decide to leave the theoretical analysis of the proposed training algorithm to future work.
\end{remark}

\begin{remark}
Before moving onto concrete examples, we briefly comment on the statistical guarantee of the $\dnase$ algorithm. One could have established convergence rate of $\hat{\sfb} (\cdot)$ to $\sfb^{\ast} (\cdot)$ using standard arguments from $M$-estimation theory \citep{schmidt2020nonparametric, van2023weak}, and hence the asymptotic normality of $\sqrt{N} (\hat{\bbeta} - \bbeta^\ast)$. However, we decide not to take this route and only demonstrate the performance of our proposed algorithm via numerical experiments (see Section \ref{sec:sim}). This is because this proof strategy ignores the effect of the training algorithm \citep{neyshabur2017implicit, goel2020superpolynomial, vempala2019gradient}, and thus the obtained guarantee is misleading \citep{xu2022deepmed}.
\end{remark}

\begin{algorithm}
\caption{Pseudocode for $\dnase$}\label{alg:fred}
\begin{algorithmic}[1]
\STATE \textbf{main hyperparameters}: alternating frequency $\gamma$, Monte Carlo sample sizes $J_{1}$ and $J_{2}$ for approximating the outer and inner integrals of $\mathcal{L}_{K}$, convergence tolerance $\delta$, maximal iteration $M$ and other common hyperparameters for DNN such as width, depth, step size, learning rate, etc;
\STATE \textbf{input}: data samples: $\{O_i\}_{i=1}^N$; Monte Carlo samples for the outer and inner integrals of $\mathcal{L}_{K}$: $\{\mathbf{t}_l\}_{l=1}^{J_1}$, $\{\mathbf{s}_j\}_{j = 1}^{J_{2}}$;
\STATE set initial values $\bbeta_{0}$ and $\bm{\omega}_{0}$ randomly;
\FOR{$m = 1, \cdots, M$}
    \STATE fix $\bm{\omega}_{m - 1}$, update $\bbeta_{m - 1}$ to $\bbeta_{m}$ by one-step minimization of $\mathcal{L}_{\psi}$ via GD (or its variants);
    \STATE fix $\bbeta_{m}$, update $\bm{\omega}_{m - 1}$ to $\bm{\omega}_{m}$ by $\gamma$-step minimization of $\mathcal{L}_{K}$ via GD (or its variants);
    \STATE set $\hat{\bbeta} \equiv \bbeta_{m}, \hat{\bm{\omega}} \equiv \bm{\omega}_{m}$;
    \IF{$\Vert \bbeta_{m} - \bbeta_{m - 1} \Vert_2 < \delta$ and $\Vert \bm{\omega}_{m} - \bm{\omega}_{m - 1} \Vert_2 < \delta$}
    \STATE break the for loop;
    \ENDIF
\ENDFOR
\STATE \textbf{output}: $(\hat{\bbeta}, \widehat{\boldsymbol{\omega}})$.
\end{algorithmic}
\end{algorithm}


\section{Motivating Examples}
\label{sec:examples}

We now provide three concrete motivating examples to illustrate the general formulation detailed in the previous section. These three examples are drawn from, respectively, regression parameter estimation under the Missing-Not-At-Random (MNAR) missing data mechanism, sensitivity analyses in causal inference, and average loss (risk) estimation in transfer learning under dataset shift. These examples demonstrate the generality of the generic estimation problem \eqref{general form sum} in semiparametric statistics.


\subsection{A Shadow Variable Approach to MNAR}
\begin{example}
\label{mnar_data}
First, we consider the problem of parameter estimation when the response variable $Y$ is MNAR using the approach introduced in \citet{zhao2022versatile}. The observed data is comprised of $N$ independent and identically distributed (i.i.d.) observations $\{X_{i}, A_{i}, Y_{i}\}_{i = 1}^{N}$ drawn from the following DGP:
\begin{equation}
\small
\label{model-mnar}
\begin{split}
X & \sim \calN (0.5, 0.25), \\
Y | X & \sim \beta_1^\ast + \beta_2^\ast X + \calN (0, 1), \\
A | Y & \sim \text{Bern} \left( \expit (1 + Y) \right), \\
\text{and } & \text{$Y$ is missing if $A = 0$}.
\end{split}
\end{equation}
Here, the missing-indicator $A$ depends on the potentially unobserved value of $Y$. The parameter of interest is the coefficients $\bbeta = (\beta_{1}, \beta_{2})^{\top}$, while the nuisance parameter is $\bm\eta = (\p_{X}, \p_{A | Y})$. The difficulty lies in the fact that $\p_{A | Y}$ is unidentified because one never observes $Y$ when $A = 0$. \citet{zhao2022versatile} showed that the following semiparametric estimator $\hat{\bbeta}$ of $\bbeta^{\ast}$ is asymptotically unbiased by positing an arbitrary model $\tilde{\p}_{A | Y}$ for $\p_{A | Y}$:  Let $\eta_{2} (y) \coloneqq \tilde{\P} (A = 1 | Y = y)$, and $\hat{\bbeta}$ is defined as:
{\small\begin{align}
\hat{\bbeta} = & \arg\min_{\bbeta \in \bbR^{2}} \left\{ \frac{1}{N} \sum_{i = 1}^{N} \psi (X_{i}, Y_{i}, A_{i}; \bbeta, \hat{\sfb}) \right\}^2, \text{where} \nonumber \\
\hat{\sfb} = & \arg\min_{\sfb} \frac{1}{N} \sum_{i=1}^N \int_{t} \Bigg( \int \sfb (s,X_i) K (s,t,X_i; \hat{\bbeta}) \diff s \nonumber \\
& \quad \quad -C (t, X_i; \hat{\bbeta}) - \p_{Y | X} (t,X_i; \hat{\bbeta}) \sfb (t,X_i) \Bigg)^2 \diff \nu (t). \label{mnar_eq1}
\end{align}}
Here, we have
\begin{align*}
& \ \psi (X_{i}, Y_{i}, A_{i}; \bbeta, \sfb) \\
\equiv & \ \frac{1}{N} \sum_{i = 1}^{N} \Bigg\{ 
 A_i \frac{\frac{\partial}{\partial \bbeta} \p_{Y | X} (Y_i, X_{i}; \bbeta)}{\p_{Y | X} (Y_i, X_{i}; \bbeta)} - A_{i} \sfb (Y_{i}, X_{i})  \\ 
- & \ (1-A_i)\frac{\int \frac{\partial}{\partial \bbeta} \p_{Y | X} (Y_i, X_{i}; \bbeta) \eta_{2} (t) \diff t}{\int \p_{Y | X} (t, X_i; \bbeta) (1 - \eta_{2} (t)) \diff t} \\
 + & \ (1 - A_{i}) \frac{\int \p_{Y | X} (t, X_{i}; \bbeta) \sfb (t, X_{i}) \eta_{2} (t) \diff t}{\int \p_{Y | X} (t, X; \bbeta) (1 - \eta_{2} (t)) \diff t}  \Bigg\},
\end{align*}
and $C (t, X_i; \bbeta)$ and $K (s, t, X_i; \bbeta)$ are known functions up to $\bbeta$ defined as follows:
\begin{equation*}
\small
\begin{split}
& C (t,X_i; \bbeta) \equiv \frac{\partial}{\partial \bbeta} \p_{Y | X} (t,X_i; \bbeta) \\ 
  & + \frac {\int \frac{\partial}{\partial \bbeta} \p_{Y | X} (s, X_i; \bbeta) \eta_{2} (s) \diff s} {\int \p_{Y | X} (s, X_i; \bbeta) (1 - \eta_{2} (s)) \diff s} \p_{Y | X} (t,X_i; \bbeta),\\
& K (s,t, X_i; \bbeta) \equiv \frac{\p_{Y | X} (s, X_i; \bbeta) \p_{Y | X} (t,X_i; \bbeta) \eta_{2} (s)}{\int \p_{Y | X} (s', X_i; \bbeta) (1 - \eta_{2} (s')) \diff s'}.
\end{split}
\end{equation*}
The nuisance parameter $\eta_{1} \equiv \p_{X}$ is ancillary to the parameter of interest $\bbeta$, so it does not appear in the above estimation procedure. The proposed estimator $\hat{\bbeta}$ in \eqref{mnar_eq1} in fact solves the so-called ``semiparametric efficient score equation'' \citep{tsiatis2007semiparametric}, the semiparametric analogue of the score equation in classical parametric statistics. Derivations of the above estimator can also be found in Appendix \ref{app:proofmnar} for the sake of completeness.

\label{eg:mnar}
\end{example}

\subsection{Sensitivity Analysis in Causal Inference}

\begin{example}
The second example is about sensitivity analysis in causal inference with unmeasured confounding \citep{robins2000sensitivity}. Specifically, we consider the DGP below with $\bfX$ of higher-dimension than that of \citet{zhang2022semi}:
\begin{equation}
\small
\label{model-sens}
\begin{split}
& \bfX \coloneqq (X_{1}, \cdots, X_{10})^{\top} \mathop{\sim}\limits^{\text{i.i.d.}} \text{Uniform} (0, 1), \\
& U^{\dag} \sim X_1 - X_2 ^2 + \calN (0, 0.1), \\
& A | \bfX, U^{\dag} \sim \text{Bern} \left( \expit \left\{ 3 \sum_{j=1}^{10} (-1)^{j+1}X_{j} + 2 U^{\dag} \right\} \right), \\
& Y | \bfX, A, U^{\dag} \sim \text{Bern} \left( \expit \left\{ 4 \sum_{j=1}^{10}(-1)^{j+1}X_{j} + \beta^\ast A + 2 U^{\dag} \right\} \right).
\end{split}
\end{equation}
Here the parameter of interest $\beta \in \mathbb{R}$ with true value $\beta^\ast$ encodes the causal effect of $A$ on $Y$. Since $U^{\dag}$ is unmeasured, $\beta$ is unidentifiable. Sensitivity analyses shall be conducted to evaluate the potential bias resulted from some fictitious $U$ specified by the statistician. \citet{zhang2022semi} first posit an arbitrary working model for (one of the nuisance parameters) $\tilde{\p} (U | \bfX) \coloneqq \eta (U, \bfX)$ inducing a possibly misspecified working joint model $\tilde{\p}$, and then define the following estimator $\hat{\beta}$ of $\beta^{\ast}$ that minimizes the loss as follows:
\begin{equation}
\begin{split}
\hat{\beta} = & \arg\min_{\beta \in \bbR} \left\{ \frac{1}{N} \sum_{i = 1}^{N} \psi (\bfX_{i}, A_{i}, Y_{i}; \beta, \hat{\sfb}) \right\}^2, \text{where}\\
\hat{\sfb} = & \arg\min_{\sfb} \sum_{i=1}^N \int_{t} \Bigg(\int \mathsf{b} (u', \bfX) K (u', u, \bfX; \hat{\beta}) \diff u' \\
& \quad \quad - C (u, \bfX; \hat{\beta}) + \lambda \mathsf{b} (u, \bfX) \Bigg)^2 \diff \nu(u),
\end{split}
\label{sen_eq1}
\end{equation}
where
\begin{equation*}
\small
\begin{aligned}
& \psi (\bfX_{i}, A_{i}, Y_{i}; \beta,\sfb) \\
& = \frac{\int f(Y_i,A_i,\bfX_i,u;\beta) \p_{Y, A | \bfX, U} (Y_{i}, A_{i}| \bfX_{i}, u; \beta) \eta (u, \bfX_{i}) \diff u}{\int \p_{Y, A | \bfX, U} (Y_{i}, A_{i}| \bfX_{i}, u; \beta) \eta (u, \bfX_{i}) \diff u}, \\
&  f (Y_i,A_i,\bfX_i,u;\beta) \coloneqq \tilde{\sfs}_{\beta} (Y_{i}, A_{i}, \bfX_{i}, u;\beta) - \mathsf{b} (u, \bfX_{i}).
\end{aligned}
\end{equation*}
Here $\tilde{\sfs}_{\beta} (y, a, \bfX, u)$ is the score of the working joint model $\tilde{\p}$ with respect to $\beta$, and $C (u, \bfX; \beta)$ and $K (u', u, \bfX; \beta)$ are known functions up to unknown parameters of the forms:
\begin{equation*}
\small
\begin{split}
& C (u, \bfX;\beta)  =  \int \frac{\int \tilde{\sfs}_{\beta} (y, a, \bfX, u') h (y, a, u', \bfX; \beta) \diff u'}{g(y, a, \bfX;\beta)} \\
& \quad \quad \quad \quad \quad \quad \quad \ \times \p_{Y, A | \bfX, U} (y, a;\beta | \bfX, u) \diff y \diff a ,\\
& K (u', u, \bfX;\beta)  = \int \frac{\p_{Y, A | \bfX, U} (y, a | \bfX, u'; \beta) }{g (y, a, \bfX;\beta)} \\
& \quad \quad \quad \quad \quad \quad \quad \ \times \p_{Y, A | \bfX, U} (y, a | \bfX, u; \beta) \eta (u', \bfX) \diff y \diff a,
\end{split}
\end{equation*}
where
\begin{equation*}
\begin{split}
& h (y, a, u', \bfX;\beta)  = \p_{Y, A | \bfX, U} (y, a| \bfX, u'; \beta) \eta (u', \bfX), \\
& g(y, a, \bfX;\beta)  = \int \p_{Y, A | \bfX, U} (y, a | \bfX, u'; \beta) \eta (u', \bfX) \diff u'.
\end{split}
\end{equation*}

Derivations of the estimator \eqref{sen_eq1} can also be found in Appendix \ref{app:Proof_sens} for the sake of completeness.
\label{eg:sensitivity}
\end{example}

\subsection{Transfer Learning: Average Loss Estimation under Dataset Shift}
\begin{example}
One of the goals of transfer learning \citep{zhang2020domain, gong2016domain, scott2019generalized, li2023subspace} is to improve estimation and inference of the parameter of interest in the target population by borrowing information from different but similar source populations. Here we consider the scenario where the target and source populations differ under the so-called ``Covariate Shift Posterior Drift'' (CSPD) mechanism introduced in \citet{qiu2023efficient}, building upon \citet{scott2019generalized}. We consider the following DGP restricted by the CSPD mechanism (see Section 6 of \citet{qiu2023efficient}). The data consists of covariates $X$, outcome $Y$, and a binary data source indicator $A$ with $A = 0/1$ for the target/source population. Let $\phi (x)$ be any function with non-zero first derivative and the data is generated as follows:
\begin{equation}
\begin{aligned}
\label{model_data_shift}
& X \sim \text{Uniform}(1,3), A | X = x \sim \text{Bern} (\expit \circ \pi (x)) \\
& Y | X = x, A = a \sim \text{Bern} \left( \expit \circ r_{a} (x) \right) \ \text{where } \\
& \pi (x) \coloneqq \logit \ \P (A = 1 | X = x), \\
& r_{a} (x) \coloneqq \logit \ \P (Y = 1 | X = x, A = a)\text{ and }\\
&  r_{1} (x) \equiv \phi \left( r_{0} (x) \right).
\end{aligned}
\end{equation}
The nuisance parameter is $\bm{\eta} = (\p_X, \p_{A|X},\p_{Y|A,X})$, with $\hat{\bm{\eta}} = (\hat{\p}_X,\hat{\p}_{A|X},\hat{\p}_{Y|A,X})$ denoting their estimates. In the procedure shown later, many quantities depend on $\bm{\eta}$ and we also attach $\hat{\cdot}$ to denote their estimates. $\bm{\eta}$ and those related quantities can be estimated by possibly nonparametric methods including DNN \citep{farrell2021deep, xu2022deepmed, chen2024causal} with sample splitting \citep{chernozhukov2018double}, but this is tangential to the main message of this paper. The parameter of interest is the average squared error loss under the target population: $\beta^{\ast} \coloneqq \E [\ell(X,Y) | A = 0] \equiv \E [(Y - f(X))^2 | A = 0]$, where $f (x)$ is any prediction function for $Y$ possibly outputted from some machine learning algorithm. \citet{qiu2023efficient} proposed the following statistically optimal (i.e. semiparametric efficient) estimator $\hat{\beta}$ of $\beta^{\ast}$:
\begin{equation}
\small
\begin{split}
& \hat{\beta} \coloneqq \frac{1}{N} \sum_{i=1}^N \Bigg\{ \frac{(1 - A_i) \hat{v} (X_{i})}{\hat{\P} (A = 0)} \\
& \quad + A_i \left\{Y_i - \expit \circ \phi \circ \hat{r}_{0} (X_i) \right\} \hat{g}_1(X_i)\\
& \quad +(1-A_i)\left\{Y_i - \expit \circ \hat{r}_{0} (X_i)\right\} \hat{g}_2(X_i) \Bigg\},\\
& \hat{\sfb} = \arg\min_{\sfb} \mathcal{L}_{K} (\sfb)  \text{ where } K (x) \coloneqq \frac{\hat{h} (x, 1)}{\hat{\mu} (x)} \hat{\p}_{X} (x), \\
& \mathcal{L}_{K} (\sfb) \coloneqq \int \Bigg(
\int_{x' \in \mathcal{S}(x)} K (x') \sfb(x') \diff x' + C (x) - \sfb(x) \Bigg)^2 \diff \nu(x).
\end{split}
\label{datashift_eq}
\end{equation}
We now unpack the above definition: here $v (x) \coloneqq \E [\ell (X, Y) | X = x, A = 0]$ is the average conditional squared error loss in the target population with $\hat{v}$ being its estimate, $\mathcal{S}(x) \coloneqq \{x': \hat{r}_{0} (x') = \hat{r}_{0} (x)\}$ is a subset in the sample space of the covariate $X$, and further define
{\small\begin{align*}
& \hat{h}(x, a) \coloneqq\hat{\var} (Y | A=a, X=x) \hat{\P}(A=a | X=x), \\
& \hat{C}(x) \coloneqq\frac{\hat{h}(x , 0) \int_{x' \in \mathcal{S} (x)} \hat{h}(x', 1) \hat{\p}_{X} (x') \diff x'}{\hat{h}(x, 1)\left\{\phi^{\prime} \circ \hat{r}_{0}(x)\right\}^2} \frac{\ell(x, 1)-\ell(x, 0)}{\hat{\P}(A=0)} ,\\
& \hat{\mu}(x) \coloneqq\int_{x' \in \mathcal{S} (x)} \hat{h}(x', 1) \hat{\p}_{X} (x') \diff x' \left\{1+\frac{\hat{h}(x , 0)}{\hat{h}(x, 1) \phi^{\prime} \circ \hat{r}_{0}(x)^2}\right\}.
\end{align*}}
We finally turn to the definitions of $\hat{g}_{1}$ and $\hat{g}_{2}$: first let $\mathscr{A} \sfb (x) \coloneqq \int_{x' \in \mathcal{S}(x)} K (x') \mathsf{b}(x') \diff x'$, and eventually define
\begin{equation*}
\small
\begin{split}
& \hat{g}_1(x) \coloneqq -\mathscr{A} \hat{\sfb} (x) \cdot \phi^{\prime} \circ \hat{r}_{0} (x) / \mathscr{A} \hat{\mu}(x) + \phi^{\prime} \circ \hat{r}_{0} (x) \cdot \hat{\sfb} (x) / \hat{\mu}(x), \\
& \hat{g}_2(x) \coloneqq \hat{\sfb} (x) / \hat{\mu}(x).
\end{split}
\end{equation*}
Derivations of the above estimator can also be found in Appendix \ref{app:Proof_transfer} for the sake of completeness. Though not stated as such, $\hat{\beta}$ in \eqref{datashift_eq} can also be written as a minimizer to a loss function, but with a closed form.


\label{eg:shift}
\end{example}

\section{Numerical Experiments}
\label{sec:sim}
In this section, we conduct three simulation studies for the three examples given in Section \ref{sec:examples} to demonstrate the empirical performance of our proposed $\dnase$ algorithm.

\subsection{Hyperparameter Tuning}
\label{sec:Tuning}

In Section \ref{sec:examples}, we present three examples that use semiparametric models for estimation and inference that involves solving integral equations. In the first two examples, our objective is to estimate the regression parameters/causal effects. In the third example, we aim to estimate the risk (average loss) function in transfer learning problems under dataset shift. The hyperparameters of the proposed $\dnase$ model include: the alternating frequency $\gamma$, Monte Carlo sizes $J_1$ and $J_2$ for numerically approximating the integrals for the integral equations and the related loss $\mathcal{L}_{K}$, and other common DNN tuning parameters such as depth, width, step sizes, batch sizes, and etc. These hyperparameters are tuned via grid search using cross-validation. We employ the hyperbolic tangent ($\tanh$) function as the activation function \citep{lu2022machine}, and set the learning rate to be $10^{-4}$. Detailed information regarding hyperparameter tuning can be found in Appendix \ref{app:hyper}.

\subsection{Simulation Results}
\subsubsection{Example \ref{eg:mnar}: Regression Parameter Estimation under MNAR}
\label{sec:mnar}
First, we apply the $\dnase$ algorithm to the problem of  parameter estimation in regression models with the response variable $Y$ being subject to MNAR as described in \eqref{model-mnar} in Example \ref{eg:mnar}.  Here we are interested in estimating the regression coefficient $\bbeta = (\beta_{1}, \beta_{2})^{\top}$, the true value of which is set to be $\bbeta^{\ast} = (0.25, -0.5)^{\top}$ in the simulation. The sample size $N$ is 500.

We choose $\eta_{2} (y) \equiv \tilde{\P} (A = 1 | Y = y) = \expit (1 - y)$ which is completely misspecified compared to the truth, repeat the simulation experiments for 100 times, and numerical summary statistics of all estimators are given in Table \ref{table:simu_1} and boxplots in Figure \ref{fig:simu_1} in Appendix \ref{app:table}. We compare the estimation errors $\hat{\bbeta} - \bbeta^\ast$ over 100 simulated datasets of the $\dnase$ algorithm against polynomials with degrees varying from two to five, together with the estimates from \citet{zhao2022versatile} using the algorithm of \citet{atkinson1967numerical} (Table \ref{table:simu_1}). From Table \ref{table:simu_1}, it is evident that the estimator $\hat{\bbeta}$ from $\dnase$ has smaller bias than polynomials. Our result has similar bias to that of \citet{zhao2022versatile}, while our variance is smaller.

\begin{table}[htbp]
\caption{This table displays the mean and standard error of $\hat{\beta}_1-\beta^{\ast}_1$ and $\hat{\beta}_2-\beta^{\ast}_2$ over 100 simulations. Our method is displayed as $\dnase$, compared to the other methods using $d$-th degree polynomial, with $d \in \{2,3,4,5\}$, shown as quadratic, cubic, quartic and quintic in the table.}
\label{table:simu_1}
\begin{center}
\begin{small}
\begin{sc}
\vskip 0.1in
\begin{tabular}{ccc}
\toprule
Mean(std)    & $\dnase$  & \citet{zhao2022versatile}  \\
\midrule
$\hat{\beta}_1 - \beta_1^\ast$ & 0.012(0.142) & -0.016(0.209)   \\ 
$\hat{\beta}_2 - \beta_2^\ast$  & 0.004(0.104) & 0.004(0.122)  \\
\toprule
Mean(std)  &   quintic & quartic \\
\midrule
$\hat{\beta}_1 - \beta_1^\ast$  & -0.113(0.070)& -0.081(0.186)   \\ 
$\hat{\beta}_2 - \beta_2^\ast$  & 0.182(0.095) & 0.124(0.112)   \\
\toprule
Mean(std)    &  cubic  & quadratic  \\
\midrule
$\hat{\beta}_1 - \beta_1^\ast$ & -0.143(0.133) & -0.033(0.178)   \\ 
$\hat{\beta}_2 - \beta_2^\ast$   & 0.130(0.158) & 0.092(0.124)  \\
\bottomrule
\end{tabular}
\end{sc}
\end{small}
\end{center}
\end{table}

\subsubsection{Example \ref{eg:sensitivity}: Sensitivity Analysis in Causal Inference}
\label{sec:sensitivity}

Next, we apply the $\dnase$ algorithm to the problem of sensitivity analyses in causal inference with unmeasured confounding described in \eqref{model-sens} in Example \ref{eg:sensitivity}. Recall in this problem we are interested in estimating the causal effect parameter $\beta$ under user-specified fictitious unmeasured confounder $U$ and sensitivity analyses models. We set the true parameter value $\beta^{\ast} \equiv 2$. The sample size $N$ is 1000.

We choose $\eta (u, \mathbf{x}) = \mathbbm{1} \{u \in [-0.5, 0.5]\}$ (defined in Example \ref{eg:sensitivity}) which is completely misspecified compared to the true $\p_{U | \bfX} (u | \bfx)$. Table \ref{table:simu_2} displays the numerical estimator of $\hat{\beta} - \beta^{\ast}$ over 100 simulated datasets obtained from various methods. It is evident that our proposed $\dnase$ algorithm is less biased and more efficient compared to polynomials with varying degrees. We also compared $\dnase$ with a grid-based method adopted in \citet{zhang2022semi} using their R code; the latter method has longer computation time (about 35 minutes per run) than $\dnase$ (below 30 minutes per run) but larger bias and variance. Table \ref{table:simu_2} provides the summary statistics of the simulation results, supplemented with boxplots in Figure \ref{fig:simu_sens} in Appendix \ref{app:table}.

\begin{table}[htbp]
\caption{The table displays the mean and standard error $\hat{\beta}-\beta^{\ast}$ over 100 simulations. Our method is displayed as $\dnase$, compared to the other methods including a grid-based method used in \citet{zhang2022semi} and $d$-th degree polynomial, with $d \in \{2, 3\}$, shown as quadratic, cubic in the table.}
\label{table:simu_2}
\begin{center}
\begin{small}
\begin{sc}
\vskip 0.1in
\begin{tabular}{ccc}
\toprule
Mean(std)    & $\dnase$  & grid-based method \\
\midrule
$\hat{\beta} - \beta^\ast$ & 0.036(0.155) & 0.109(0.426)  \\ 
\toprule
Mean(std)    & cubic & quadratic  \\
\midrule
$\hat{\beta} - \beta^\ast$ &  0.142(0.189) & 0.145(0.188)   \\ 
\bottomrule
\end{tabular}
\end{sc}
\end{small}
\end{center}
\end{table}

\subsubsection{Example \ref{eg:shift}: Average Loss Estimation in Transfer Learning}

Next we apply $\dnase$ to the transfer learning problem described in \eqref{model_data_shift} in Example \ref{eg:shift}. Recall in this problem we are interested in estimating the average target-population loss $\beta^{\ast}$ given in Example \ref{eg:shift}. Here we simply choose $f (x) \equiv \expit (x^{\frac{1}{2}})$ as the prediction function for $Y$, $\phi (x) \equiv x + 1$ as the outcome shift mechanism function, and $r_{a} (x) \equiv \pi (x) \equiv x$ as the logit-transformed outcome and data-source conditional expectations. Note that to focus on the main issue of this paper, in the simulation we use the true nuisance parameter $\bm{\eta}$ and related quantities such as $v, h, \cdots$ defined in Example \ref{eg:shift}, without using their estimates. The sample size $N$ is 10000.

The numerical summary statistics of the discrepancies between the estimated parameters, $\hat{\beta}$, and the true parameters, $\beta^\ast$, across 100 simulated datasets are presented in Table \ref{table:simu_3}. The corresponding boxplots, which provide a visual representation of these results, have been relegated to Figure \ref{fig:shift}. It is evident that our proposed $\dnase$ method outperforms the traditional polynomial method with different degrees, in particular in terms of variance (reduced by at least 80\%).
\vskip -0.1in
\begin{table}[htbp]
\caption{This table displays $\hat{\beta} - \beta^\ast$ over 100 simulations. Our method is displayed as $\dnase$, compared to methods using $d$-th degree polynomials, with $d \in \{5, 10, 15\}$, respectively displayed as ``quintic'', ``10th-degree'', ``15th-degree'' in the table.}
\label{table:simu_3}
\begin{center}
\begin{small}
\begin{sc}
\vskip 0.1in
\begin{tabular}{ccc}
\toprule
Mean(std)    & $\dnase$  &   quintic  \\
\midrule
$\hat{\beta} - \beta^*$ & -0.0008(0.0084) & 0.0009(0.0186)  \\ 
\toprule
Mean(std)    & 10th-degree &  15th-degree \\
\midrule
$\hat{\beta} - \beta^\ast$ &  0.0033(0.0234) & 0.0038(0.0385)  \\
\bottomrule
\end{tabular}
\end{sc}
\end{small}
\end{center}
\end{table}

\section{Real Data Application}
\label{sec:data}

In this section, we further demonstrate the empirical performance of $\dnase$ by reanalyzing a real dataset that was previously analyzed in \citet{zhao2022versatile}.

\subsection{The Connecticut Children's Mental Health Study}

We now apply our proposed $\dnase$ algorithm to re-analyze a dataset from a comprehensive investigation conducted by \citet{zahner1997factors} on children's mental health in Connecticut, USA. This study employed the Achenbach Child Behavior Checklist (CBCL) as a tool to evaluate the psychiatric status of children. The CBCL consists of three different forms, namely \textit{parent}, \textit{teacher}, and \textit{adolescent self-report}. Here the response $Y$ of interest is teacher's report of the child's psychiatric status, where $Y = 1$ suggests the presence of clinical psychopathology and $Y = 0$ suggests normal psychological functioning. In the study, a logistic regression was conducted to examine the relationship between $Y$ and covariates, including $X_{1}$: father's presence in the household ($X_{1} = 1/0$: absence/presence), $X_{2}$: the child's physical health ($X_{2} = 1/0$: poor/good health), $Z$: parent's report ($Z = 1/0$: abnormal/normal clinical psychopathology):
\begin{align*}
\small
\logit \ \P (Y = 1 | Z, X_{1}, X_{2}) = \beta_{0} + \beta_{1} X_{1} + \beta_{2} X_{2} + \beta_{z} Z,
\end{align*}
where the regression coefficients $\bbeta = (\beta_{0}, \beta_{1}, \beta_{2}, \beta_{z})^{\top}$ are the parameters of interest. 

This dataset comprises 2486 subjects, of which 1061 (approximately 42.7\%) exhibit missing values. The presence of such a large amount of missing data poses a significant challenge, as modeling with only the 1425 subjects with complete data may lead to highly biased estimates for $\bbeta$. The missingness $A$ of teacher's report $Y$ is unlikely to be related to the parent's report $Z$, but $Z$ may be highly correlated with $Y$. Hence $Z$ can be viewed as a so-called ``shadow variable'' \citep{miao2024identification} to help identify $\bbeta$ even under MNAR. The semiparametric estimator of $\bbeta$ leveraging the shadow variable has been derived in \citet{zhao2022versatile}. A more detailed exposition can be found in Appendix \ref{app:data} .

\subsection{Data Analysis Results}

Besides the estimator computed by $\dnase$, for comparison purposes, we also report the results (1) from the observed-only estimator that completely ignores the missing data and (2) from \citet{zhao2022versatile} using the algorithm of \citet{atkinson1967numerical}. The observed-only estimator is biased under the MNAR assumption for the regression coefficients of the covariates except $Z$, the shadow variable.

To estimate the standard errors of the $\dnase$ estimators of $\bbeta$, we generate 100 bootstrap samples, each of size 2000, rerun the whole $\dnase$ algorithm to obtain the point estimates, and then compute the mean and standard deviation of the point estimates over 100 bootstrap samples. Table \ref{real_data_table} displays the point estimates and the estimated standard errors from different methods, including $\dnase$, the reported estimates from Table 6 of \citet{zhao2022versatile}, and the potentially biased observed-only estimates. First, it is reassuring that all three methods produce roughly the same estimate for regression coefficient $\beta_{z}$ of the shadow variable $Z$ (parent's report), although our method, $\dnase$, has the smallest standard error (the 2nd row, 2nd column of Table \ref{real_data_table}). For the other coefficients, the point estimates of $\dnase$ are very close to those from \citet{zhao2022versatile} but our (estimated) standard errors of $\dnase$ are smaller. Overall, the results of $\dnase$ are consistent with that of \citet{zhao2022versatile}. The ``observed-only'' approach is more likely to produce biased result due to its stronger yet empirically unverifiable assumption on the missing data mechanism.
\begin{table}[htbp]
\setlength\tabcolsep{3pt}
\small
    \centering
    \caption{Comparison of different estimators for $\bbeta$ in the real data analysis.}
    \vskip 0.15in
    \begin{sc}
    \begin{tabular}{ccc}
        \toprule
        Mean(se) &  $\hat{\beta}_{0}$ &  $\hat{\beta}_{1}$ \\
        \midrule
        $\dnase$ & -1.3289(0.1402) & -0.0623(0.1196)  \\
        \citet{zhao2022versatile} & -1.3585(0.1823) & -0.0718(0.1470)\\
        observed-only & -1.9307(0.1132) & 0.3652(0.1480) \\
        \midrule
        Mean(se) & $\hat{\beta}_{2}$ & $\hat{\beta}_{z}$ \\
        \midrule
        $\dnase$ & -0.6159(0.1366) & 1.4620(\textbf{0.0761}) \\
        \citet{zhao2022versatile} & -0.9817(0.1320) & 1.4623(\textbf{0.1194}) \\
        observed-only & -0.0516(0.1690) & 1.4621(\textbf{0.1583}) \\
        \bottomrule
    \end{tabular}
    \end{sc}\label{real_data_table}
\end{table}
\section{Concluding Remarks}
\label{sec:conclude}

In this paper, we first proposed a general framework by formulating the semiparametric estimation problem as a bi-level optimization problem; and then we developed a novel algorithm $\dnase$ that can simultaneously solve the integral equation and estimate the parameter of interest by employing modern DNN as a numerical solver of the integral equations appeared in the estimation procedure. Compared to traditional approaches, $\dnase$ can be scaled to higher-dimensional problems owing to: (1) the universal approximation property \citep{siegel2020approximation}; (2) more efficient optimization algorithms \citep{kingma2015adam}; and  (3) the increasing computing power for deep learning. We also developed a python package that implements the $\dnase$ algorithm. In fact, this proposed framework goes beyond semiparametric estimation and can be adapted more broadly to models subject to general moment restrictions in the statistics and econometrics literature \citep{ai2003efficient, chen2018overidentification}.

There are some recent works introducing modern numerical methods, such as methods based on gradient flows \citep{crucinio2022solving} or tensor networks (e.g. tensor train decomposition) \citep{cichocki2016tensor, cichocki2017tensor, corona2017tensor, chen2023learninggan}, to the problem of purely solving integral equations. It will be interesting to explore how to incorporate these new generations of numerical toolboxes into semiparametric statistics. 
Our future plan also includes: (1) providing convergence analysis of Algorithm \ref{alg:fred} and possibly supplementing it with consensus-based methods \citep{carrillo2021consensus} to accelerate the convergence towards global optimizer \textit{in practice}; (2) characterizing statistical properties of the obtained estimator; and more importantly, (3) fully automating and ``democratizing'' semiparametric statistics by eliminating the need of deriving estimators by human, via techniques such as large language models, arithmetic formula learning, differentiable and symbolic computation \citep{frangakis2015deductive, carone2019toward, polu2020generative, garg2020learning, trinh2024solving, luedtke2024simplifying}. 



\section*{Acknowledgements}
The authors would like to thank \href{https://sites.google.com/site/blsunnus/}{BaoLuo Sun} for helpful discussions. Lin Liu is supported by NSFC Grant No.12101397 and No.12090024, Shanghai Science and Technology Commission Grant No.21ZR1431000 and No.21JC1402900, and Shanghai Municipal Science and Technology Major Project No.2021SHZDZX0102. Zixin Wang and Lei Zhang are also supported by Shanghai Science and Technology Commission Grant No.21ZR1431000. Lin Liu is also affiliated with Shanghai Artificial Intelligence Laboratory and the Smart Justice Lab of Koguan Law School of SJTU. Zhonghua Liu is also affiliated with Columbia University Data Science Institute.

\section*{Impact Statement}

The goal of this paper is to advance the field of Machine Learning/Artificial Intelligence by using these tools to boost statistical sciences. There could be many potential positive societal consequences of our work, none of which we feel must be specifically highlighted here.
 
\bibliography{example_paper}
\bibliographystyle{icml2024}

\newpage
\appendix
\onecolumn
\appendix
\section{The flow chart of $\dnase$}
\label{app:figs}

\begin{figure}[htbp]
    \centering
    \includegraphics[width = 1.0\textwidth]{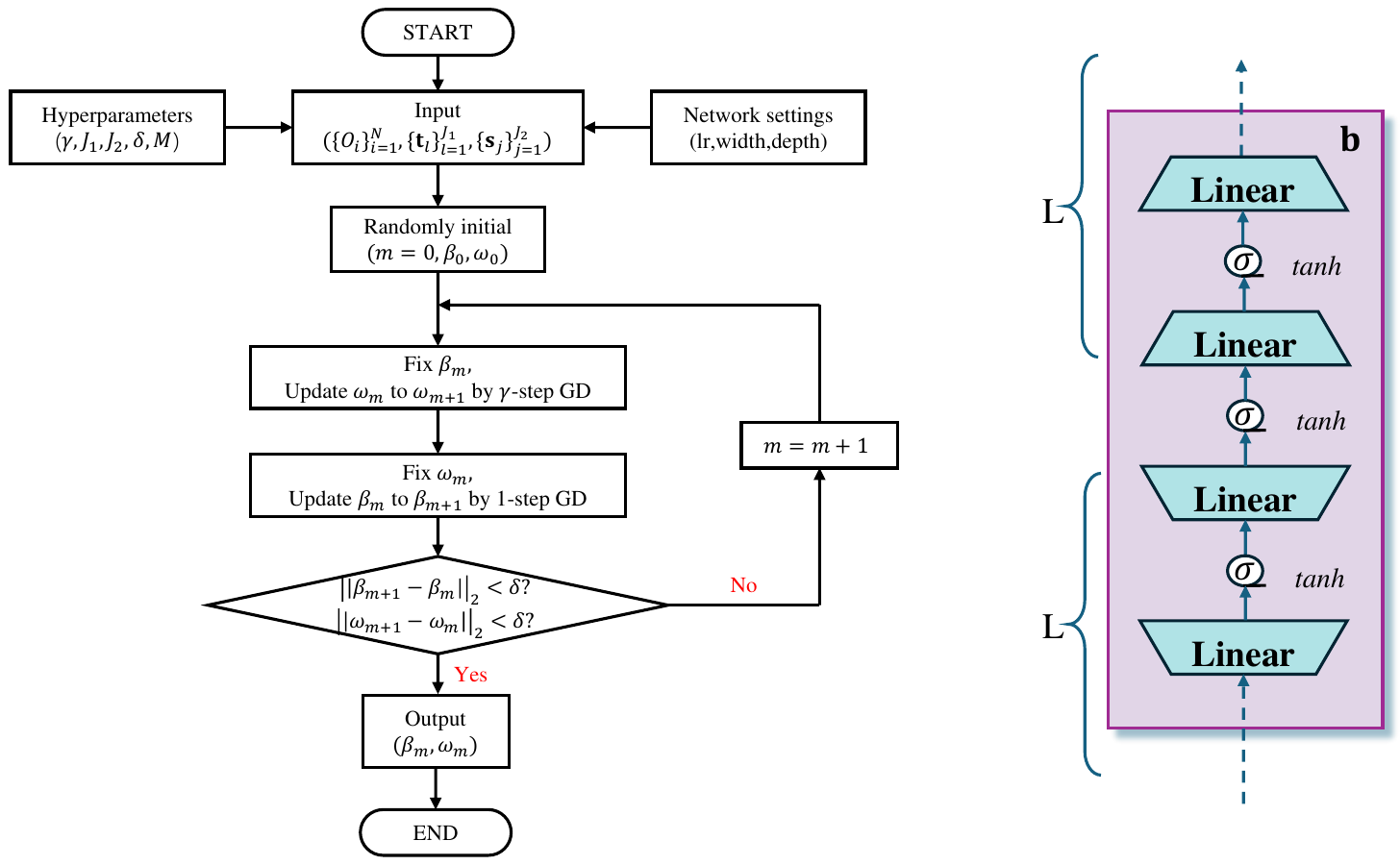}
    \caption{Depicted here is an illustration of our $\dnase$ algorithm and the neural network architecture employed for modeling $\sfb$, the solution to the integral equations. Left: We present a comprehensive flowchart that outlines the algorithmic procedure. Right: The architecture of $\sfb$ is chosen to be the standard Multilayer Perceptron (MLP) or feedforward neural networks in this paper with tanh activation function. Notably, the MLP is constructed with a depth of $2L+1$, indicating a strategic layering to optimize performance.}
    \label{fig:fred}
\end{figure}

\section{Further remarks on the loss function \eqref{loss_func}}
\label{app:loss}

As mentioned in Section \ref{sec:loss}, the (empirical) loss function \eqref{loss_func} originates from the following population loss function:
\begin{equation}
\label{pop loss}
\begin{split}
& \bbeta^{\ast} \coloneqq \arg\min_{\bbeta \in \bbR^{q}} \mathcal{L}_{\psi} (\bbeta, \sfb^{\ast}) \ \st \ \sfb^{\ast} (\cdot) \coloneqq \arg\min_{\sfb \in \calF} \mathcal{L}_{K} (\sfb, \bbeta^\ast), \\
& \text{where } \mathcal{L}_{\psi} (\bbeta, \sfb) \coloneqq \left\{ \E \left[ \psi (O; \sfb (\cdot), \bbeta) \right] \right\}^{2}, \\
& \mathcal{L}_{K} (\sfb) \coloneqq \int_{o} \int_{\mathbf{t}} \left( \int K(\mathbf{s}, \mathbf{t}, o; \bbeta) \mathsf{b} (\mathbf{s}, o) \diff \mathbf{s} - C(\mathbf{t}, o); \bbeta)-\sfb (\mathbf{t}, o) \right)^2 \diff \nu (\mathbf{t}) \diff \P (o).
\end{split}
\end{equation}
We first turn the estimating equation \eqref{general form sum} into the above population loss and then replace the integral with respect to $\diff \P (o)$ by the empirical average.

We would like to make another remark on this loss function. One can in principle first draw many samples from the space where the parameter of interest $\bbeta$ lies, and then solve the integral equations over all these possible values. However, the Alternating GD type of algorithms can explore the $\bbeta$ space more efficiently.

Finally, it is noteworthy that we are interested in the global minimizer of the loss \eqref{pop loss}. Since the $\dnase$ algorithm relies on GD, when multiple local minimizers exist, one could in principle consider a trial-and-error approach and initialize multiple starting points for $\bbeta$ and the function $\sfb(\cdot)$ to ensure convergence to the global minimizer. One could also use the idea from consensus-based optimization \citep{carrillo2021consensus} by initializing different starting points using Interacting Particle Systems to accelerate the trial-and-error approach by injecting benign correlations among those starting points.

\section{Derivations for Section \ref{sec:examples}}

To make our paper relatively self-contained, we provide derivations of the estimators for the examples in Section \ref{sec:examples}. We want to remark that these derivations are not new and have appeared in the cited works in Section \ref{sec:examples}.

\subsection{Derivations for Example \ref{eg:mnar}}
\label{app:proofmnar}
From the model given by Example \ref{mnar_data}, the true missing-data mechanism is $\P (A = 1 | Y = y) = \expit (1 + y)$. Given a dataset consisting of $N$ i.i.d. observations $(X_i, A_i, A_i Y_i)^N_{i=1}$, the observed-data joint likelihood and the observed-data score are given below:
\begin{align*}
    & \p (x, a, y; \bbeta, \eta) = \p_X (x) \left\{\p_{Y | X} (y, x; \bbeta) \P (A = 1 | Y = y) \right\}^a \left\{1-\int \p_{Y | X} (t, x; \bbeta) \P (A = 1 | Y = t) \mathrm{d} t \right\}^{1 - a}, \\
& \sfs_\bbeta(x, a, a y; \bbeta) = a \frac{\frac{\partial}{\partial \beta} \p_{Y | X}(y, x ; \bbeta)}{\p_{Y | X} (y, x ; \bbeta)} - (1 - a) \frac{\int \frac{\partial}{\partial \bbeta} \p_{Y | X}(t, x ; \bbeta) \P (A = 1 | Y = t) \mathrm{d} t}{1-\int \p_{Y | X}(t, x; \bbeta) \P (A = 1 | Y = t) \mathrm{d} t},
\end{align*}
where $\p_{Y | X} (y, x; \bbeta)$ is assumed to be known up to the parameter value $\bbeta$, but $\eta$ is completely unknown to the statistician. From semiparametric theory, the nuisance tangent space is $\Lambda=\Lambda_1 \oplus \Lambda_2$, where
\begin{align*}
\Lambda_{1} = \{h(x): \E h (X) = 0\}, \Lambda_2 & =\left\{a g (y)-(1-a) \frac{\int \p_{Y | X}(t, x ; \bbeta) g (t) \eta(t) \mathrm{d} t}{1-\int \p_{Y | X}(t, x ; \bbeta) \eta(t) \mathrm{d} t}: \forall \ g (\cdot)\right\}.
\end{align*}
By definition, one can obtain the observed-data efficient score $\sfs_{\eff, \bbeta} (x, a, ay)$ with respect to $\bbeta$ by subtracting from $\sfs_{\bbeta} (x, a, ay)$ its projection onto the observed-data nuisance tangent space $\Lambda$. The projection operation leads to the terms involving function $\sfb (\cdot)$ (corresponding to the function $g$ in $\Lambda_2$) in the display below:
\begin{equation*}
    \sfs_{\eff, \bbeta}(x, a, a y)=\sfs_\bbeta(x, a, a y, \bbeta)-a \sfb(y)+(1-a) \frac{\int \p_{Y \mid X}(t, x; \bbeta) \sfb(t ; \bbeta) \pi(t) \mathrm{d} t}{1-\int \p_{Y \mid X}(t, x ; \bbeta) \pi(t) \mathrm{d} t},
\end{equation*}
where $\sfb(y)$ satisfies the following function:
\begin{equation*}
    \begin{aligned}
& \int\left\{\frac{\partial}{\partial \bbeta} \p_{Y \mid X}(y, x ; \bbeta)+\frac{\int \frac{\partial}{\partial \bbeta} \p_{Y \mid X}(t, x ; \beta) \eta(t) \mathrm{d} t}{\int \p_{Y \mid X}(t, x ; \bbeta)(1-\eta(t)) \diff t} \p_{Y \mid X}(y, x ; \bbeta)\right\} \p_X(x) \mathrm{d} x \\
= & \int\left\{\sfb(y ; \bbeta) \p_{Y \mid X}(y, x ; \bbeta)+\frac{\int \sfb(t ; \bbeta) \p_{Y \mid X}(t, x ; \bbeta) \eta(t) \mathrm{d} t}{\int \p_{Y \mid X}(t, x ; \bbeta)(1-\eta(t)) \mathrm{d} t} \p_{Y \mid X}(y, x ; \bbeta)\right\} \p_X(x) \mathrm{d} x.
\end{aligned}
\end{equation*}
\citet{zhao2022versatile} showed that a working model $\eta (y)$ can be used to replace the true missing mechanism $\P (A = 1 | Y = y)$, which eventually leads to the estimator $\hat{\bbeta}$ that solves the following empirical version of the estimating equation:
\begin{equation*}
\begin{aligned}
& \ \frac{1}{N} \sum_{i=1}^N \Bigg\{\sfs_\bbeta \left(X_i, A_i, A_i Y_i\right)-A_i \sfb\left(Y_i ; \bbeta\right)+\left(1-A_i\right) \frac{\int \p_{Y \mid X}\left(t, X_i ; \bbeta\right) \sfb(t ; \bbeta) \eta(t) \mathrm{d} t}{\int \p_{Y \mid X}(t, X ; \bbeta)\left(1-\eta(t)\right) \mathrm{d} t}\Bigg\}=0, \\
 \text{where} & \\
& \ \frac{1}{N} \sum_{i=1}^N\left\{\frac{\partial}{\partial \bbeta} \p_{Y \mid X}\left(y, X_i ; \bbeta\right)+\frac{\int \frac{\partial}{\partial \bbeta} \p_{Y \mid X}\left(t, X_i ; \bbeta\right) \eta (t) \mathrm{d} t}{\int \p_{Y \mid X}\left(t, X_i ; \bbeta\right)\left(1-\eta(t)\right) \mathrm{d} t} \p_{Y \mid X}\left(y, X_i ; \bbeta\right)\right\} \\
= & \ \frac{1}{N} \sum_{i=1}^N\left\{\sfb (y ; \bbeta) \p_{Y \mid X}\left(y, X_i ; \bbeta\right)+\frac{\int \sfb(t ; \bbeta) \p_{Y \mid X}\left(t, X_i ; \bbeta\right) \eta (t) \mathrm{d} t}{\int \p_{Y \mid X}\left(t, X_i ; \bbeta\right)\left(1-\eta(t)\right) \mathrm{d} t} \p_{Y \mid X}\left(y, X_i ; \bbeta\right)\right\} .
\end{aligned}
\end{equation*}
Intuitively, this is because one eliminates the information from the nuisance parameters (in particular the guessed model for $A | Y$) in the observed-data efficient score by projecting onto the orthocomplement of the observed-data nuisance tangent space.

\subsection{Derivations for Example \ref{eg:sensitivity}}
\label{app:Proof_sens}

In this example, the likelihood of the complete data can be factorized as follows:
\begin{align*}
\begin{split}
\p (\bfX = \bfx, U = u, A =  a, Y = y)  = \p_{\bfX} (\bfx) \p_{U | \bfX} (u | \bfx) \p_{A | \bfX, U} (a | \bfx, u) \p_{Y | \bfX, U, A} (y | \bfx, u, a; \beta).
\end{split}
\end{align*}
Here the parameter of interest is $\beta$ and the remaining pieces of the joint likelihood are treated as the nuisance parameters $\bm{\eta}$. 
One can derive the observed-data score with respect to $\beta$ as the conditional expectation of the full-data score with respect to $\beta$: $\sfs_{\beta} (\bfX, A, Y) = \E \left[\sfs_\beta(\bfX, U, A, Y) | \bfX, A, Y\right]$ \citep{robins1994estimation, van2003unified}. However, one could obtain more efficient estimators which also rely on fewer modeling assumptions by calculating the observed-data efficient score based on $\sfs_{\beta} (\bfx, a, y)$ as:
$$
\sfs_{\eff, \beta}(\bfX, A, Y) \equiv \sfs_\beta (\bfX, A, Y) - \E \left[ \sfb (U, \bfX) \mid \bfX, A, Y \right],
$$
where the second term in the above display is simply the projection of $\sfs_{\beta} (\bfX, A, Y)$ onto the orthocomplement of the observed-data nuisance tangent space $\Lambda$ to be defined later, which further implies that $\sfb (u,\bfx)$ solves the following the integral equation:
$$
\E\left[\sfs_\beta(\bfX, A, Y) \mid \bfX, U\right]=\E\left\{\E [\sfb (U, \bfX) \mid \bfX, A, Y] \mid \bfX, U\right\}.
$$

To derive the observed-data nuisance tangent space, one first obtains the complete-data nuisance tangent space $\Lambda^F$:
\begin{equation*}
\begin{aligned}
& \Lambda^F=\Lambda_1^F \oplus \Lambda_2^F, \\
& \Lambda_1^F=\left\{a_1(\bfx): \E\left[a_1(\bfX)\right]=0\right\}, \\
& \Lambda_2^F=\left\{a_2(u, \bfx): \E\left[a_2(U, \bfX) \mid \bfX\right]=0\right\}.
\end{aligned}
\end{equation*}
It is well-known that the observed-data model tangent space is the expectation of the complete-data model tangent space conditioning on the observed data \citep{robins1994estimation, van2003unified}. Hence the observed-data tangent space $\Lambda$ is:
\begin{equation*}
    \begin{split}
        & \Lambda = \Lambda_1 \oplus \Lambda_2,\\
& \Lambda_1 \equiv \E [\Lambda_1^F | \bfX, A, Y] \equiv \Lambda_1^F \equiv \left\{a_1(\bfx): \E\left[a_1(\bfX)\right]=0\right\}, \\
& \Lambda_2 \equiv \E [\Lambda_2^F | \bfX, A, Y] \equiv \left\{a_2' (\bfx, a, y) \equiv \E \left[a_2(U, \bfX) \mid \bfX = \bfx, A = a, Y = y\right]: \E\left[a_2(U, \bfX) \mid X\right]=0\right\}.
    \end{split}
\end{equation*}
Then the projection of $\sfs_{\beta} (\bfX, A, Y)$ onto $\Lambda$ is simply to find the function $\sfb (U, \bfX)$ such that $\E [\sfs_{\beta} (\bfX, A, Y) -  \E [\sfb (U, \bfX) | \bfX, A, Y] | \bfX, U] \equiv 0$ (for this step, see Equation (6) in the online supplementary materials of \citet{zhang2022semi}), which is a Fredholm integral equation of the first kind, an ill-posed problem. However, one can turn this ill-posed problem to a well-posed problem by introducing Tikhonov regularization, which is the route taken in \citet{zhang2022semi}.

Since $\p_{U | \bfX} (u | \bfx)$ cannot be identified from the observed data distribution, one postulate a working model for $\p_{U | \bfX} (u | \bfx)$ as $\tilde{\p} (u| \bfx) \coloneqq \eta (u, \bfx) \mathbbm{1} \{u \in [-B, B]\} / 2 B$ where $B$ is sufficiently large. Thus, under this working model, the integral equation (after Tikhonov regularization by a possibly very small tuning parameter $\lambda$) becomes
\begin{equation*}
\int \sfb^\ast (u', \bfX) K (u', u, \bfX) \diff u' = C (u, \bfX) - \lambda \sfb^{\ast} (u, \bfX).
\end{equation*}
where $C (u, \bfX)$ and $K (u, u', \bfX)$ are known functions up to the unknown parameters. In particular, $C (u, \bfX)$ and $K (u', u, \bfX)$ are of the forms
\begin{equation*}
\begin{split}
C (u, \bfX)  = & \int \frac{\int \tilde{\sfs}_{\beta} (y, a, \bfX, u') h (y, a, u', \bfX) \diff u'}{g(y, a, \bfX)} \p_{Y, A | \bfX, U} (y, a | \bfX, u) \diff y \diff a ,\\
K (u', u, \bfX)  = & \int \frac{\p_{Y, A | \bfX, U} (y, a | \bfX, u') \p_{Y, A | \bfX, U} (y, a | \bfX, u)\eta (u', \bfX)}{g (y, a, \bfX)} \diff y \diff a ,
\end{split}
\end{equation*}
where
\begin{equation*}
\begin{split}
h (y, a, u', \bfX) & = \p_{Y, A | \bfX, U} (y, a | \bfX, u') \eta (u', \bfX), \\
g(y, a, \bfX) & = \int \p_{Y, A | \bfX, U} (y, a | \bfX, u') \eta (u', \bfX) \diff u'.
\end{split}
\end{equation*}
Eventually, one obtains the estimator $\hat{\beta}$ parameter of interest $\beta$ by solving:
\begin{equation*}
\begin{aligned}
& 0 = \frac{1}{N} \sum_{i = 1}^{N} \psi (\bfX_{i}, A_{i}, Y_{i}; \beta), \text{ where }\\
& \psi (\bfX_{i}, A_{i}, Y_{i}; \beta)  = \frac{\int f (Y_i,A_i,\bfX_i,u) \p_{Y, A | \bfX, U} (Y_{i}, A_{i} | \bfX_{i}, u) \eta (u, \bfX_{i}) \diff u}{\int \p_{Y, A | \bfX, U} (Y_{i}, A_{i} | \bfX_{i}, u) \eta (u, \bfX_{i}) \diff u},\\
& f (Y_i,A_i,\bfX_i,u) \coloneqq \tilde{\sfs}_{\beta} (Y_{i}, A_{i}, \bfX_{i}, u) - \sfb^{\ast} (u, \bfX_{i}),
\end{aligned}
\end{equation*}
where we let $\tilde{\p} (\bfx, u, a, y)$ denote the joint likelihood induced by the working model:
$$\tilde{\p} (\bfx, u, a, y) = \p_{X} (\bfx) \eta (u, \bfx) \p_{A | \bfX, U} (a | \bfx, u) \p_{Y | \bfX, U, A} (y | \bfx, u, a; \beta)$$ 
and the score for the working joint law with respect to $\beta$ is simply $\tilde{\sfs}_{\beta} (\bfX, U, A, Y) = \frac{\partial \log \tilde{\p} (\bfX, U, A, Y; \beta)}{\partial \beta}$.
\citet{zhang2022semi} showed that even when the working model is completely misspecified, the estimator based on the observed-data efficient score, not the observed-data score, is still consistent. Intuitively, this is because one eliminates the information from the nuisance parameters (in particular the guessed distribution for $U | X$) in the observed-data efficient score by projecting onto the orthocomplement of the observed-data nuisance tangent space. Proof details are provided in \citet{zhang2022semi} for the aforementioned equations.

\subsection{Derivations for Example \ref{eg:shift}}
\label{app:Proof_transfer}

In the example of target-population average loss estimation in the context of transfer learning, the main structural assumption of \citet{qiu2023efficient} is the following relationship of the logit conditional outcome models between the target ($A = 0$) and the source ($A = 1$) population: let $\phi: \bbR \rightarrow \bbR$ be a strictly increasing function with non-zero first derivative, and \citet{qiu2023efficient} impose the following:
\begin{equation}
\label{constraint}
\logit \ \P (Y = 1 | X = x, A = 1) = \phi \left(\logit \ \P (Y = 1 | X = x, A = 0) \right).
\end{equation}
As mentioned in the main text, the parameter of interest is the average squared-error loss of prediction $Y$ using a (possibly arbitrary) prediction function $f (\cdot)$ in the target population (i.e. conditioning on $A = 0$):
\begin{equation*}
\begin{split}
\beta^* & \coloneqq \E [\ell(X,Y) | A=0], \text{ where } \ell(x,y) = (y- f (x))^2.
\end{split}
\end{equation*}

The full derivation of the semiparametric efficient estimator $\hat{\beta}$ can be found in Section S9.6 of \citet{qiu2023efficient}. But we recapture the main idea here for readers' convenience. To avoid clutter, we do not repeat the definitions of the notation already appeared in Example \ref{eg:shift}. Similar to the above two examples, the derivation starts from the observed-data joint likelihood:
\begin{align*}
\p (X = x, A = a, Y = y) = & \ \p_{X} (x) \p_{A | X} (a | x) \P (Y = 1 | X = x, A = 1)^{a y} \P (Y = 0 | X = x, A = 1)^{a (1 - y)} \\
& \times \P (Y = 1 | X = x, A = 0)^{(1 - a) y} \P (Y = 0 | X = x, A = 0)^{(1 - a) (1 - y)} \\
= & \ \p_{X} (x) \p_{A | X} (a | x) \{\expit \circ \phi \circ r_{0} (x)\}^{a y} \{1 - \expit \circ \phi \circ r_{0} (x)\}^{a (1 - y)} \\
& \times \{\expit \circ r_{0} (x)\}^{(1 - a) y} \{1 - \expit \circ r_{0} (x)\}^{(1 - a) (1 - y)}.
\end{align*}
In this example, since the parameter of interest $\beta^\ast$ and the nuisance parameters are not in separate pieces of the joint likelihood decomposition, one needs to first derive one influence function $\mathsf{IF}$ of $\beta^\ast$ (a mean-zero random variable that is the functional first-order derivative of $\beta^\ast$ \citep{van2002part}), then characterize the model tangent space $\mathcal{T}$, and finally obtain the efficient influence function $\mathsf{EIF}$ of $\beta^\ast$ by projecting $\mathsf{IF}$ onto the model tangent space $\mathcal{T}$ \citep{van2003unified, tsiatis2007semiparametric}.

First, one can easily obtain one influence function of $\beta^\ast$ following standard functional differentiation argument recently reviewed in \citet{hines2022demystifying}:
\begin{align*}
\IIF = \frac{1 - A}{\P (A = 0)} (\ell (X, Y) - \beta^\ast).
\end{align*}

The model tangent space $\mathcal{T}$ under constraint \eqref{constraint} can be derived from the observed-data joint likelihood as the closure of the direct sum of the space of all scores with respect to each piece of the joint likelihood. Here since one does not model the joint likelihood parametrically, the score with respect to each piece is defined via paths of parametric submodels that go through the underlying likelihood; for details, see \citet{van2002part} or Section S9.6 of \citet{qiu2023efficient}. With constraint like \eqref{constraint}, it is often easier to derive the orthocomplement $\mathcal{T}^{\perp}$ of $\mathcal{T}$. In particular, \citet{qiu2023efficient} showed the following:
\begin{align*}
\mathcal{T}^{\perp} \equiv \left\{ \begin{array}{c}
A (Y - \expit \circ \phi \circ r_{0} (X)) f (X) - (1 - A) (Y - \expit \circ r_{0} (X)) \frac{h (X, 1) \cdot \phi' \circ r_{0} (X)}{h (X, 0)} f (X): \\
\forall \ f \ \st \ \E [h (X, 1) f (X) | r_{0} (X)] \equiv 0
\end{array} \right\}.
\end{align*}

With $\mathcal{T}^{\perp}$, one can obtain $\EIF$ by subtracting from $\IIF$ the projection of $\IIF$ onto $\mathcal{T}^{\perp}$ to obtain the final semiparametric efficient estimator $\hat{\beta}$. In particular, this projection operation in order to determine the unique $f \equiv \sfb^\ast$ in $\mathcal{T}^{\perp}$ leads to the following integral equation:
\begin{equation*}
\int _{x' \in \mathcal{S}(x)}\sfb^{\ast}(x') \frac{v(x' ; 1)}{\mu(x')} \p_{X}(x') \diff x' + C(x) = \sfb^{\ast}(x).
\end{equation*}
where $\mathcal{S}(x) = \{x' :r_{0} (x')=r_{0}(x)\}$ and the definition of $C$ can be found in Example \ref{eg:shift} (with all the $\hat{\cdot}$ removed in $\hat{C}$).
As in the main text, let:
$$
\mathscr{A} g(x)=\int _{x' \in \mathcal{S}(x)}g(x') \frac{v(x' ; 1)}{\mu(x')} \p_{X} (x') \diff x'.
$$
The efficient influence function $\EIF$ for $\beta^*$ is then
\begin{align*}
\EIF = \frac{1 - A}{\P (A = 0)} (v (X) - \beta^\ast) + A (Y - \expit \circ \phi \circ r_0 (X)) g_1 (X) + (1 - A) (Y - \expit \circ r_0 (X)) g_2 (X),
\end{align*}
with $g_1, g_2$ defined as in the main text, except removing all the $\hat{\cdot}$.

\section{Hyperparameter tuning}
\label{app:hyper}

The tuning parameters that vary the most include the width and depth of the network, as well as the alternating frequency $\gamma$. Through grid search and 5-fold cross-validation, for MNAR, the optimal network width and depth are 5 and 3, respectively, resulting in the lowest average MSE. Conversely, in the sensitivity analysis, the optimal network width and depth are 33 and 23, respectively. In the case of transfer learning, which differs from the other two examples in that the integral equations do not depend on the parameter of interest $\bbeta$, the optimal width and depth are 5 and 3. The effect of different choices of $\gamma$ is depicted in Figure \ref{sim1_fig} and \ref{sim2_fig}. In all the examples, we choose the activation function as $\tanh$, which is more commonly used in DNN for problems in scientific computing \citep{lu2022machine}. We choose the Monte Carlo sample sizes $J_1$ and $J_2$ to be 1000.

\subsection{Hyperparameter tuning for Example \ref{eg:mnar}}
\label{app:tuning_1} 

\begin{figure}[htbp]
\begin{center}
\includegraphics[width = 0.85\textwidth]{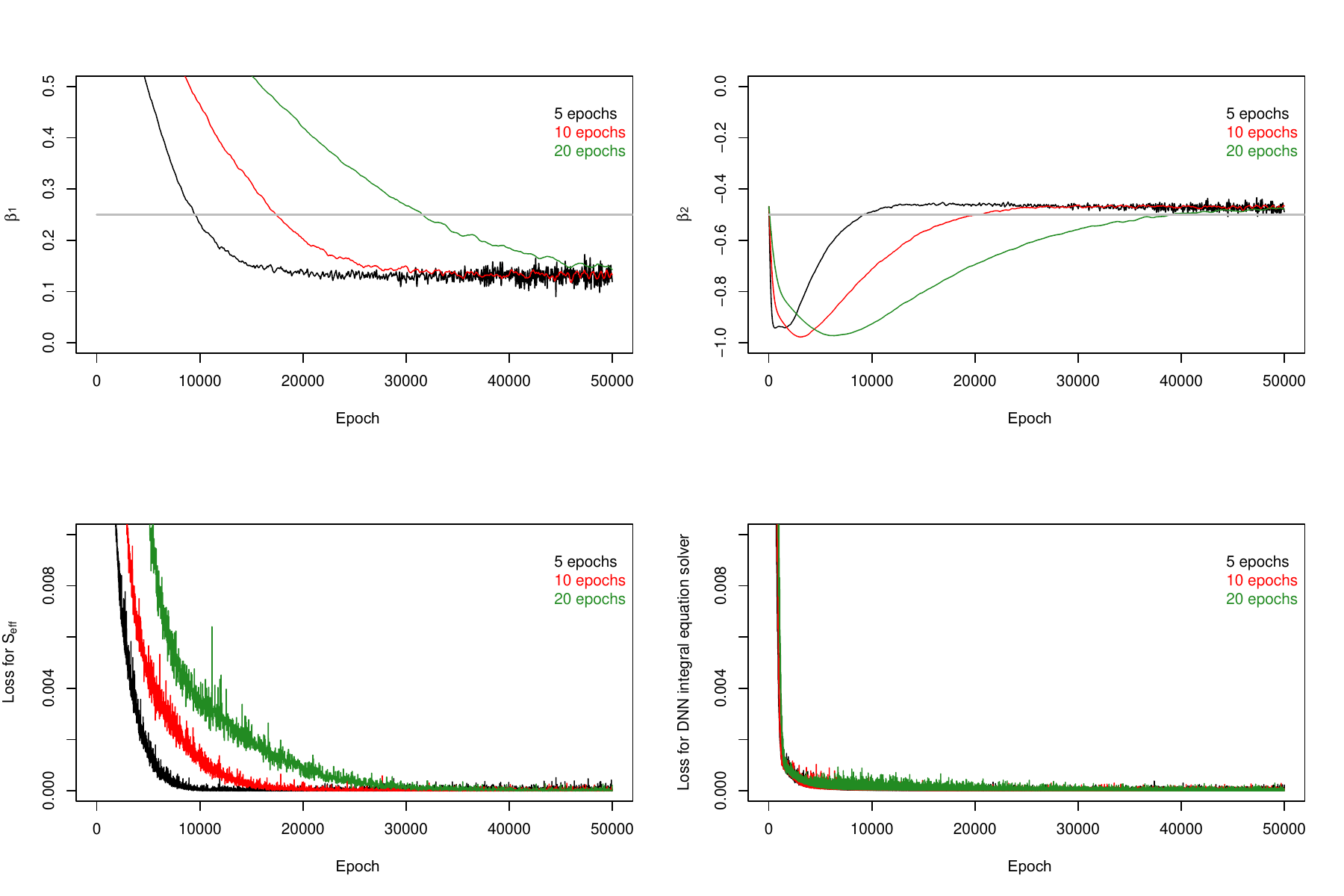}
\end{center}
\caption{The evolution of training processes in one simulated dataset of Example \ref{sec:mnar} by setting the alternating frequency $\gamma$ as $5, 10$, or $20$. The horizontal axis is the number of iterations. The upper panels show the evolution of $\beta_{1}$ (left) and $\beta_{2}$ (right) over iterations, while the lower panels show the evolution of the loss corresponding to the score equation (left) and the integral equation (right) over iterations.}
\label{sim1_fig}
\end{figure}

We find that in this example the alternating frequency $\gamma$ heavily affects the speed and stability of convergence, for the detail see Figure \ref{sim1_fig}. When $\gamma$ is set to 1, i.e. when the neuron parameters $\omega$ and the target parameter $\beta$ are updated in every other iteration, the training process fails to converge and oscillates with extremely high frequency (not shown in Figure \ref{sim1_fig}). When we increases $\gamma$, the training process starts to stabilize after a certain number of iterations and $\dnase$ eventually outputs $\hat{\bbeta} \approx \bbeta = (0.25, -0.5)^{\top}$ (Figure \ref{sim1_fig}). However, there exists a trade-off: when $\gamma = 5$, the training process converges steeply but also exhibits some instability as the iteration continues; whereas when $\gamma = 20$, it takes much longer time for the training process to converge. For the time being, we recommend practitioners try multiple $\gamma$'s (higher than 5) and check if they all eventually converge to the same values. It is an interesting research problem to study an adaptive procedure of choosing $\gamma_{\opt}$ that optimally balances speed and stability of the training. Finally, we found that the results are not sensitive to the Monte Carlo sample sizes $J_1, J_2$ as long as they are sufficiently large.

\subsection{Hyperparameter tuning for Example \ref{eg:sensitivity}}
\label{app:tuning_2}

\begin{figure}[htbp]
\centering
\begin{subfigure}{0.45\textwidth}
\begin{center}
\includegraphics[width = \textwidth]{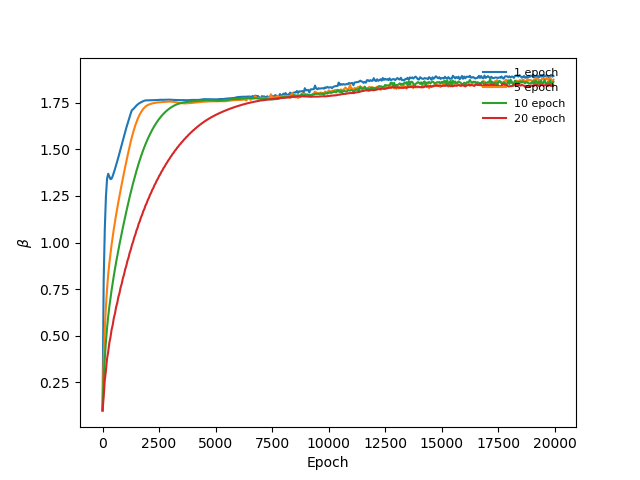}
\caption{}
\end{center}
\end{subfigure}\par\medskip
\begin{subfigure}{0.45\textwidth}
\begin{center}
\includegraphics[width = \textwidth]{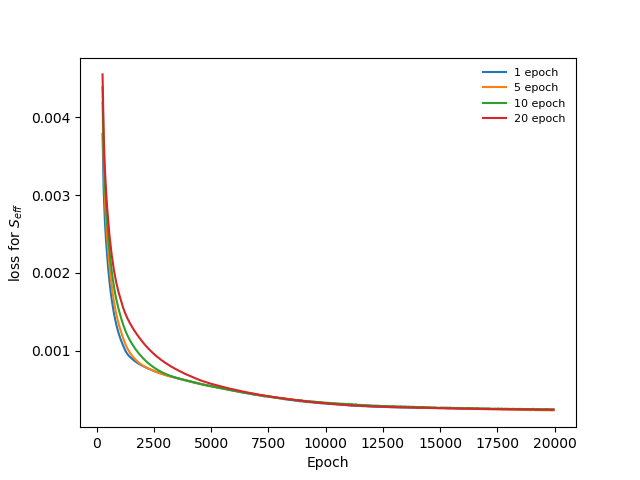}
\caption{}
\end{center}
\end{subfigure}
\begin{subfigure}{0.45\textwidth}
\begin{center}
\includegraphics[width = \textwidth]{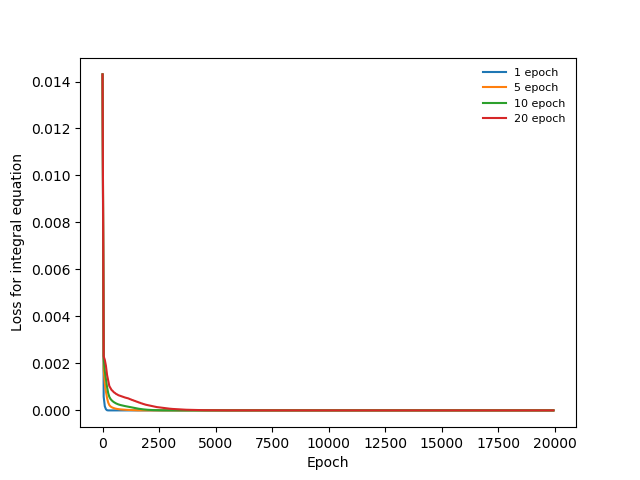}
\caption{}
\end{center}
\end{subfigure}
\caption{The evolution of training processes in one simulated dataset of Example \ref{sec:sensitivity} by setting the alternating frequency $\gamma$ as $1, 5, 10$, or $20$. The horizontal axis is the number of iterations. The upper panels show the evolution of (a) $\beta$  over iterations, while the lower panels show the evolution of the loss corresponding to (b) the score equation  and (c) the integral equation over iterations.}
\label{sim2_fig}
\end{figure}

Similar to the last example, it is imperative to examine whether various alternating frequencies will impact the outcomes. However, in contrast to Example \ref{mnar_data}, all training processes stabilize at different rates that are directly proportional to the alternating frequency. The tuning details for simulations described in Section \ref{sec:sensitivity} are presented in Figure \ref{sim2_fig}. The depicted figure provides a comprehensive analysis of the alternate frequencies $\gamma$ employed, namely 1, 5, 10, and 20, revealing their performance similarities. Notably, the convergence rate appears to be faster when $\gamma$ is set to 1, as observed from the figure. For the regularization parameter $\lambda$, we simply choose $\lambda = 0.001$ in this paper.


\newpage

\subsection{Hyperparameter tuning for Example \ref{eg:shift}}
\label{app:Tuning3}

Compared to the two previous examples, the main distinctive feature of this example is that the integral equation does not depend on the parameter of interest $\beta$. Thus in this special case, there is no need to use Alternating GD type of algorithms and regular GD type of algorithms suffices. As a result, there is no need to tune the alternating frequency $\gamma$ in Algorithm \ref{alg:fred}.

\section{Supplementary figures for Section \ref{sec:sim}}
\label{app:table}

\begin{figure}[htbp]
\centering
\begin{subfigure}{0.45\textwidth}
\begin{center}
\includegraphics[width = 1.0\textwidth]{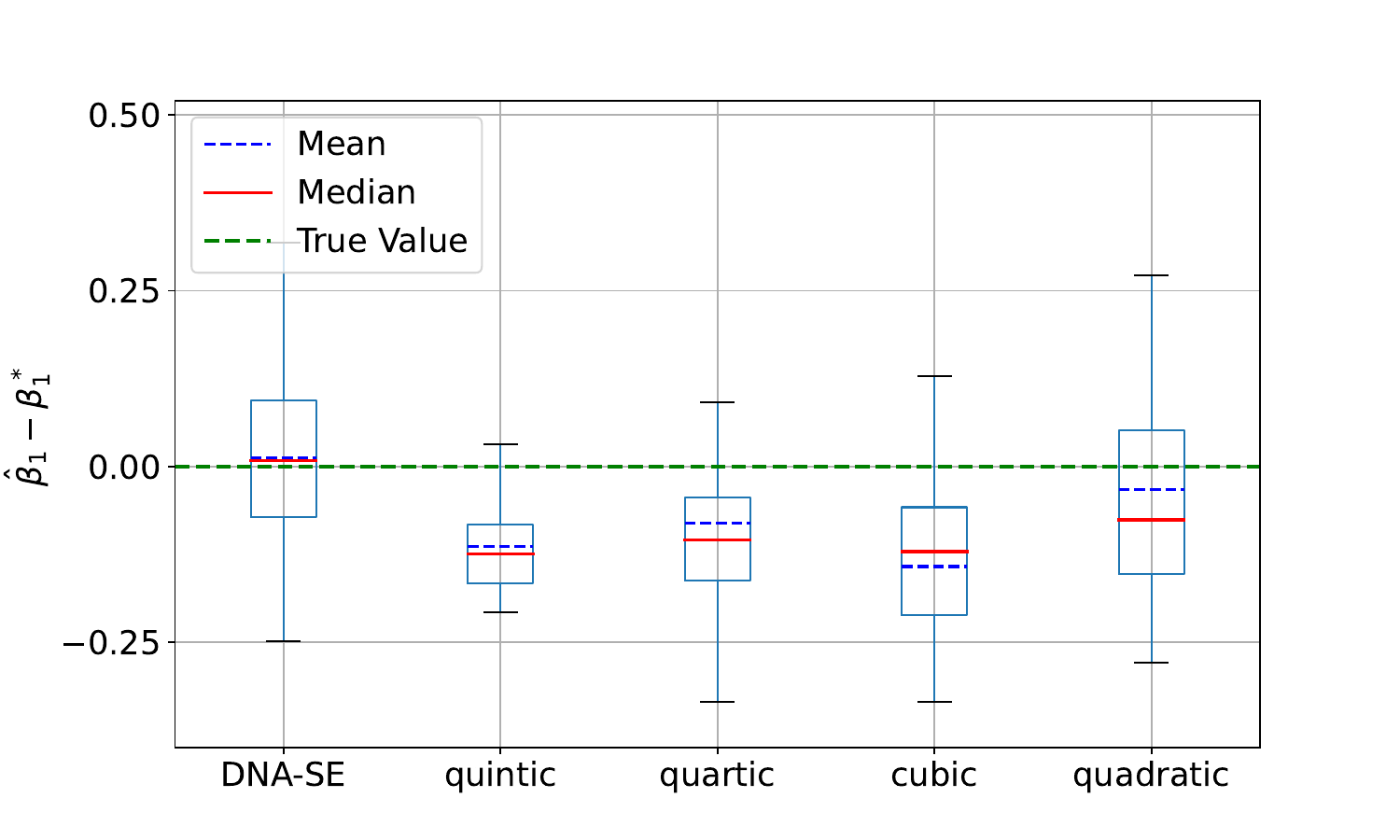}
\caption{}\label{}
\end{center}
\end{subfigure}
\begin{subfigure}{0.45\textwidth}
    \begin{center}
        \includegraphics[width = 1.0\textwidth]{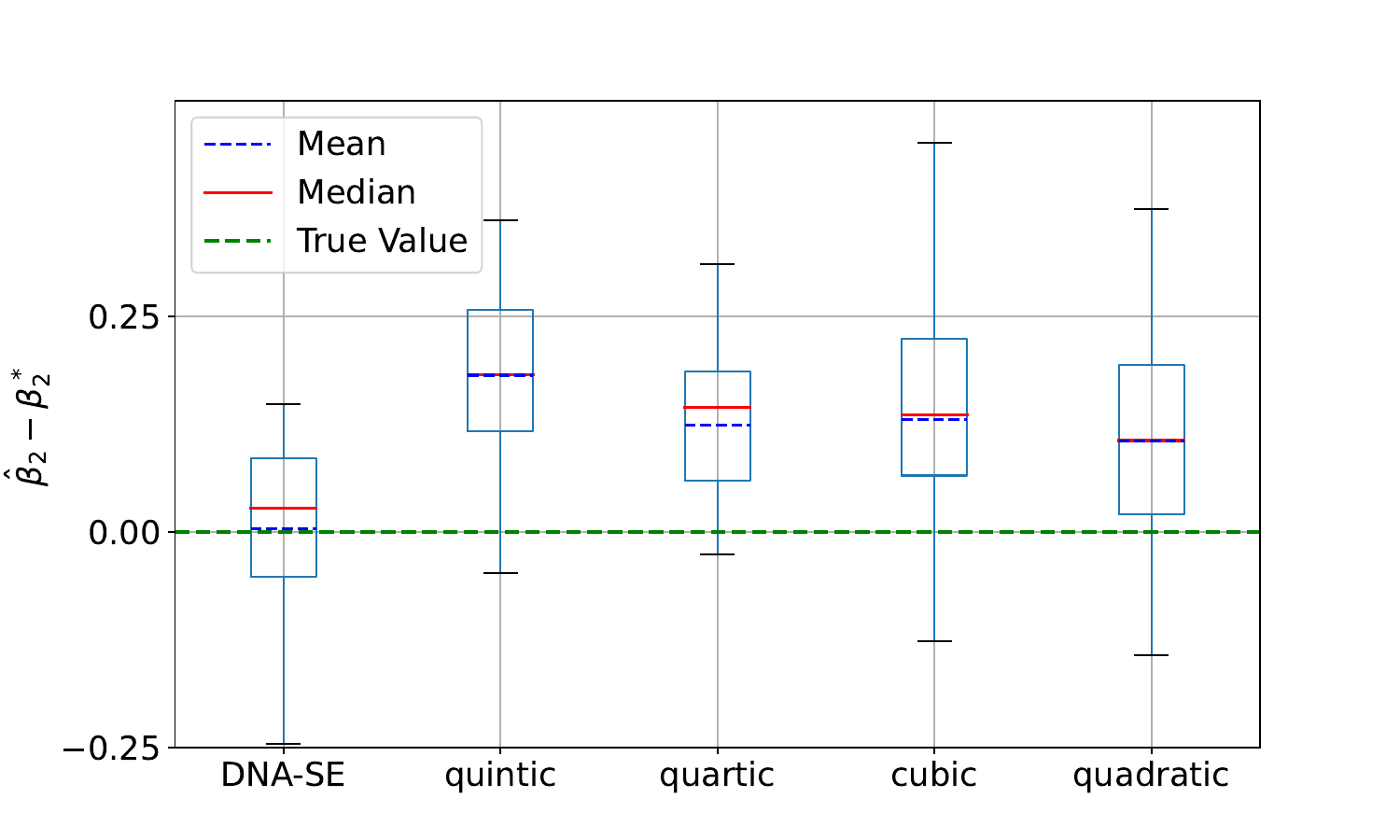}
        \caption{}
    \end{center}
\end{subfigure}
\caption{The boxplots of (a) and (b) display $\hat{\beta}_{1} - \beta_1$ and $\hat{\beta}_{2} - \beta_2$ over 100 simulations. Our method is displayed as $\dnase$, compared to the methods using $d$-th degree polynomials, with $d \in \{2, 3, 4, 5\}$, respectively displayed as ``quadratic'', ``cubic'', ``quartic'', and ``quintic'' on the horizontal axis of the plot.}
\label{fig:simu_1}
\end{figure}

\begin{figure}[htbp]
\centering
\begin{center}
\includegraphics[width = 0.5\textwidth]{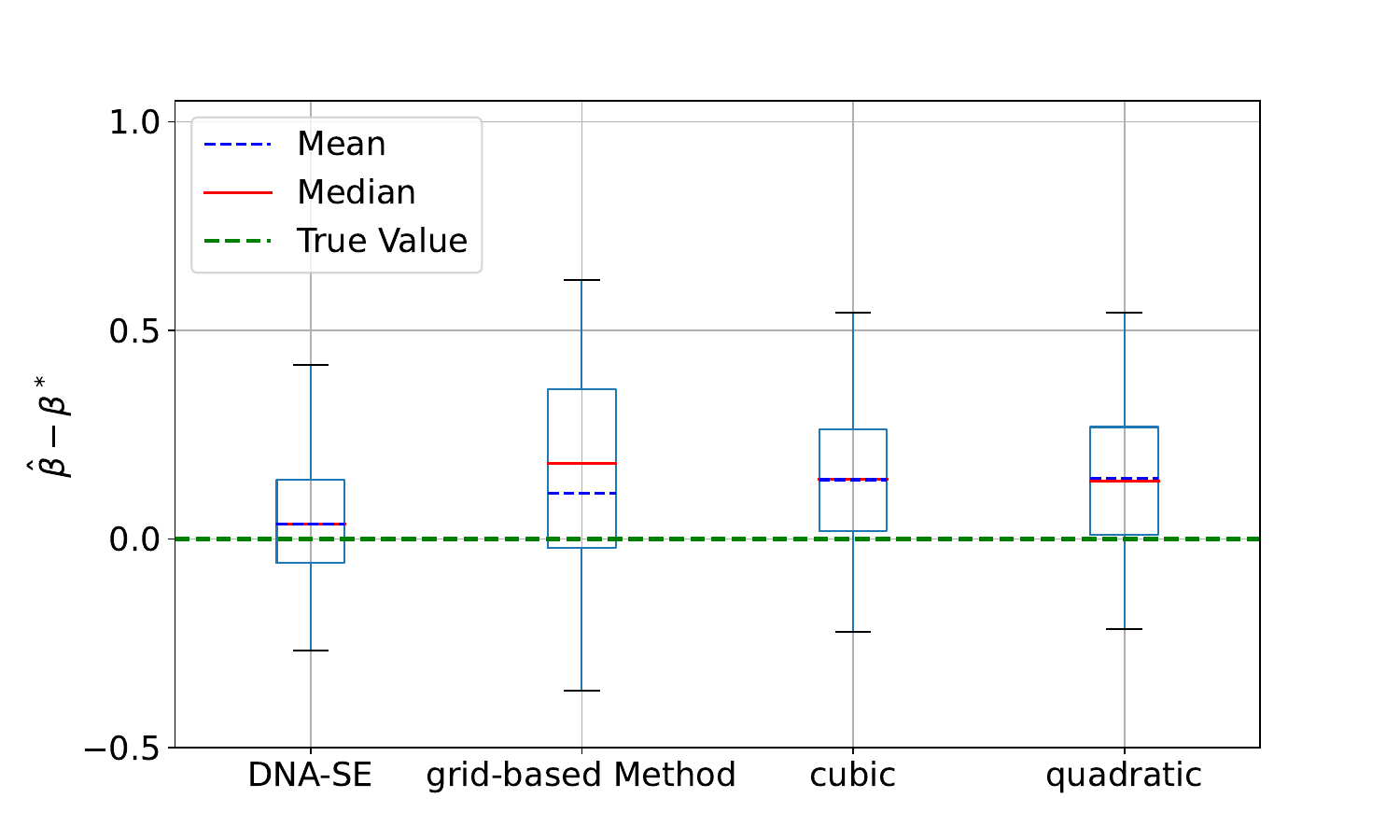}
\end{center}
\caption{The boxplot displays $\hat{\beta} - \beta$ over 100 simulations. Our method is displayed as $\dnase$, compared to the grid-based method of \citet{zhang2022semi} and the methods using $d$-th degree polynomials, with $d \in \{2, 3\}$, respectively displayed as ``quadratic'' and ``cubic'' on the horizontal axis of the plot.}
\label{fig:simu_sens}
\end{figure}

This section collects supplementary figures (Figures \ref{fig:simu_1} to \ref{fig:simu_sens}) for the numerical experiments that were only referenced in Section \ref{sec:sim} due to space limitation.

\begin{figure}[htbp]
    \centering
    \includegraphics[width = 0.5\textwidth]{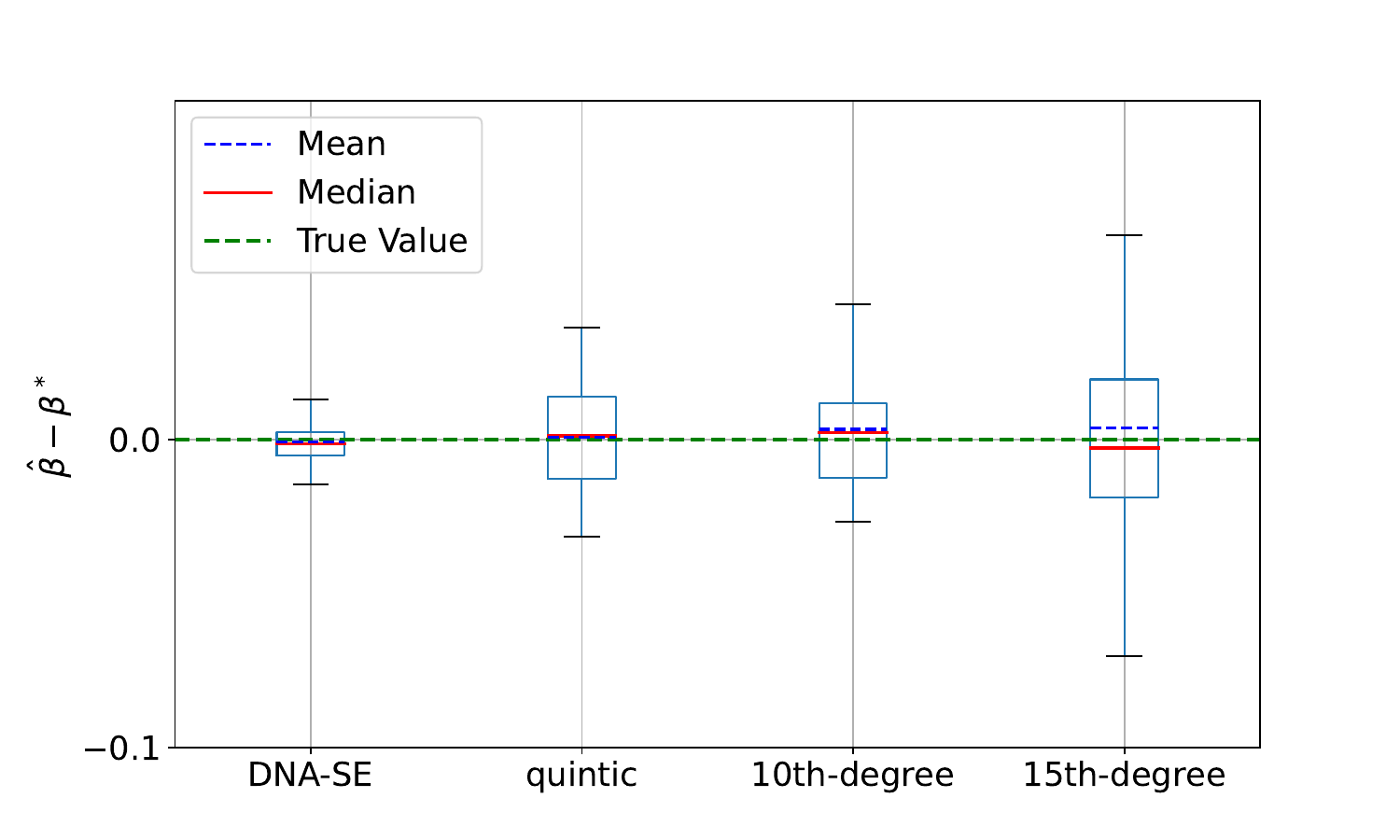}
    \caption{The boxplot displays $\hat{\beta} - \beta^\ast$ over 100 simulations. Our method is displayed as $\dnase$, compared to methods using $d$-th degree polynomials, with $d \in \{5, 10, 15\}$, respectively displayed as ``quintic'', ``10th-degree'', ``15th-degree'' on the horizontal axis of the plot.}
    \label{fig:shift}
\end{figure}

\section{The estimators used for real data analysis}
\label{app:data}

Following \citet{zhao2022versatile}, the missing data mechanism is posited to be $\eta^{\ast} (Y, X_{1}, X_{2})$, as a function of the response \textit{teacher's report}, covariates $X_{1}$ (father's presence) and $X_{2}$ (child's health):
\begin{equation*}
\begin{split}
& \ \eta^{\ast} (Y, X_{1}, X_{2}) = \logit \ \P (A = 1 | Y, X_{1}, X_{2}) = 1.058 - 2.037 Y -0.002 X_{1} + 0.298 X_{2}.
\end{split}
\end{equation*}
\citet{zhao2022versatile} also posited the following model for the shadow variable $Z$, as:
\begin{equation*}
\begin{split}
& \ \logit \ \P(Z = 1| X_{1}, X_{2}) = -2.106 + 0.623 X_{1} + 0.890 X_{2}.
\end{split}
\end{equation*}
The parameters of interest are $\bbeta = (\beta_{0}, \beta_{1}, \beta_{2}, \beta_{z})^{\top}$ and the estimator $\hat{\bbeta}$ is given below:
\begin{equation*}
\begin{split}
\hat{\bbeta} = & \arg\min_{\bbeta \in \bbR^{2}} \left\{ \frac{1}{N} \sum_{i = 1}^{N} \psi (\bfX_{i},Z_i, Y_{i}, A_{i}; \bbeta, \hat{\sfb}) \right\}^2, \text{where} \\
\hat{\sfb} = & \arg\min_{\sfb} \frac{1}{N} \sum_{i=1}^N \int_{t} \Bigg( \int \sfb (s,\bfx) K (s,t,\bfx,\bfX_i, Z_i; \hat{\bbeta}) \diff s -C (t, \bfx,Z_i; \hat{\bbeta}) - \p_{Y | \bfX, Z} (t | \bfx,Z_i; \hat{\bbeta}) \sfb (t,\bfx) \Bigg)^2 \diff \nu (t).
\end{split}
\end{equation*}
Here,
\begin{equation*}
\begin{aligned}
& \psi(\bfX_{i},Z_i, A_i, Y_i, \boldsymbol{\beta}, \sfb)  = \frac{1}{N} \sum_{i = 1}^{N} \Bigg\{ 
 A_i \frac{\frac{\partial}{\partial \bbeta} \p_{Y | \bfX, Z} (Y_i | \bfX_{i},Z_i; \bbeta)}{\p_{Y | \bfX, Z} (Y_i | \bfX_{i},Z_i; \bbeta)} - A_{i} \sfb (Y_{i},\bfX_{i}) \\ 
& - (1-A_i)\frac{\int \frac{\partial}{\partial \bbeta} \p_{Y | \bfX, Z} (Y_i | \bfX_{i},Z_i; \bbeta) \eta^{\ast} (t) \diff t}{\int \p_{Y | \bfX, Z} (t | \bfX_{i},Z_i; \bbeta) (1 - \eta^{\ast} (t)) \diff t}  +  \ (1 - A_{i}) \frac{\int \p_{Y | \bfX, Z} (t | \bfX_{i},Z_i; \bbeta) \sfb (t,\bfX_{i}) \eta^{\ast} (t) \diff t}{\int \p_{Y | \bfX, Z} (t | \bfX_{i}, Z_i; \bbeta) (1 - \eta^{\ast} (t)) \diff t} \Bigg\},
\end{aligned}
\end{equation*}
and $C(\cdot)$ and $K(\cdot)$ are defined as follows:
\begin{equation*}
\label{app:key_2}
\begin{split}
& C (t,\bfx,Z_i; \bbeta) \equiv \frac{\partial}{\partial \bbeta} \p_{Y | \bfX, Z} (t | \bfx,Z_i; \bbeta)  + \frac {\int \frac{\partial}{\partial \bbeta} \p_{Y | \bfX, Z} (s | \bfx,Z_i; \bbeta) \eta^{\ast} (s) \diff s} {\int \p_{Y | X} (s,\bfx,Z_i; \bbeta) (1 - \eta^{\ast} (s)) \diff s} \p_{Y | X} (t,\bfx,Z_i; \bbeta),\\
& K (s,t,\bfx,\bfX_i, Z_i; \bbeta) \equiv \frac{\p_{Y | \bfX, Z} (s | \bfx,Z_i; \bbeta) \p_{Y | \bfX, Z} (t | \bfx,Z_i; \bbeta) \eta^{\ast} (s)}{\int \p_{Y | \bfX, Z} (s' | \bfx,Z_i; \bbeta) (1 - \eta^{\ast} (s')) \diff s'} \text{Kernel}_{h} (\bfX_i-\bfx).
\end{split}
\end{equation*}
Here $\text{Kernel}_{h} (\cdot) \coloneqq \text{Kernel} (\cdot / h) / h$ and the kernel function is chosen based on the criteria outlined in Section 4 of \citet{zhao2022versatile}. In the simulation, we choose the same kernel function as in \citet{zhao2022versatile} with the same bandwidth $h = 3 N^{-1/5}$:
\begin{equation*}
\text{Kernel} (t)=\frac{45}{32}\left(1-\frac{7}{3} t^2\right)\left(1-t^2\right) I_{\{|t| \leq 1\}}.
\end{equation*}

\end{document}